\documentclass[journal]{IEEEtran}
\usepackage{cite}
\usepackage{graphicx}
\usepackage{bm}
\usepackage{epsfig}
\usepackage{amsmath}
\usepackage{array}
\usepackage{multirow}
\usepackage{amssymb}
\usepackage{color}
\ifCLASSINFOpdf
\else
\fi
\begin{document}
%
\title{Batch Coherence-Driven Network for Part-aware Person Re-Identification}

%
\author{Kan Wang,~\IEEEmembership{Student Member,~IEEE,}
        Pengfei Wang,
        Changxing Ding,~\IEEEmembership{Member,~IEEE,} \\
        and Dacheng Tao,~\IEEEmembership{Fellow,~IEEE}

\thanks{Manuscript received X, X; revised X, X; accepted X, X. Date of publication X, X; date of current version X, X. This work was supported by the National Natural Science Foundation of China under Grant 61702193, 62076101, and U1801262, the Program for Guangdong Introducing Innovative and Entrepreneurial Teams under Grant 2017ZT07X183, the Natural Science Fund of Guangdong Province under Grant 2018A030313869, the Science and Technology Program of Guangzhou under Grant 201804010272, the Guangzhou Key Laboratory of Body Data Science under Grant 201605030011, and the Fundamental Research Funds for the Central Universities of China under Grant 2019JQ01. The associate editor coordinating the review of this manuscript and approving it for publication was Prof. Daniel Lau. \emph{(Corresponding author: Changxing Ding.)}}

\thanks{Kan Wang and Pengfei Wang are with the School of Electronic and Information Engineering, South China University of Technology,
381 Wushan Road, Tianhe District, Guangzhou 510000, China (e-mail: eekan.wang@mail.scut.edu.cn; mswangpengfei@mail.scut.edu.cn).}
\thanks{Changxing Ding is with the School of Electronic and Information Engineering, South China University of Technology,
381 Wushan Road, Tianhe District, Guangzhou 510000, China. He is also with the Pazhou Lab, Guangzhou 510330, China (e-mail: chxding@scut.edu.cn).}
\thanks{Dacheng Tao is with JD Explore Academy at JD.com, Beijing, China (e-mail: dacheng.tao@jd.com).}
\thanks{This paper has supplementary downloadable material available at http://ieeexplore.ieee.org., provided by the author.}}

%
%

\markboth{IEEE TRANSACTIONS ON IMAGE PROCESSING, VOL. X, NO. X, XXX 2020}
{Shell \MakeLowercase{\textit{et al.}}:}
%



\maketitle
\begin{abstract}
    Existing part-aware person re-identification methods typically employ two separate steps: namely, body part detection and part-level feature extraction. However, part detection introduces an additional computational cost and is inherently challenging for low-quality images. Accordingly, in this work, we propose a simple framework named Batch Coherence-Driven Network (BCD-Net) that bypasses body part detection during both the training and testing phases while still learning semantically aligned part features. Our key observation is that the statistics in a batch of images are stable, and therefore that batch-level constraints are robust. First, we introduce a batch coherence-guided channel attention (BCCA) module that highlights the relevant channels for each respective part from the output of a deep backbone model. We investigate channel-part correspondence using a batch of training images, then impose a novel batch-level supervision signal that helps BCCA to identify part-relevant channels. Second, the mean position of a body part is robust and consequently coherent between batches throughout the training process. Accordingly, we introduce a pair of regularization terms based on the semantic consistency between batches. The first term regularizes the high responses of BCD-Net for each part on one batch in order to constrain it within a predefined area, while the second encourages the aggregate of BCD-Net's responses for all parts covering the entire human body. The above constraints guide BCD-Net to learn diverse, complementary, and semantically aligned part-level features. Extensive experimental results demonstrate that BCD-Net consistently achieves state-of-the-art performance on four large-scale ReID benchmarks.
\end{abstract}


\section{Introduction}
    \IEEEPARstart{P}{erson} re-identification (ReID) aims at retrieving pedestrian images belonging to the same identity across non-overlapping camera views. Due to its broad range of potential applications (e.g., video surveillance \cite{CVPRristani2018features}, searching for missing children, etc.), there has been explosive growth in ReID research in recent years\cite{CVPRyang2019patch, TIPyang2018metric, ICCVChen2019Mixed, ICCVGuo2019Beyond, ICCVZhou2019Omni, ICCVDai2019Batch, TIPNguyen2019kernel, TIPdai2019video, TIPfeng2020, TIPzhang2020, TIPchen2020}.

    \begin{figure}
    \centerline{\includegraphics[width=0.475\textwidth]{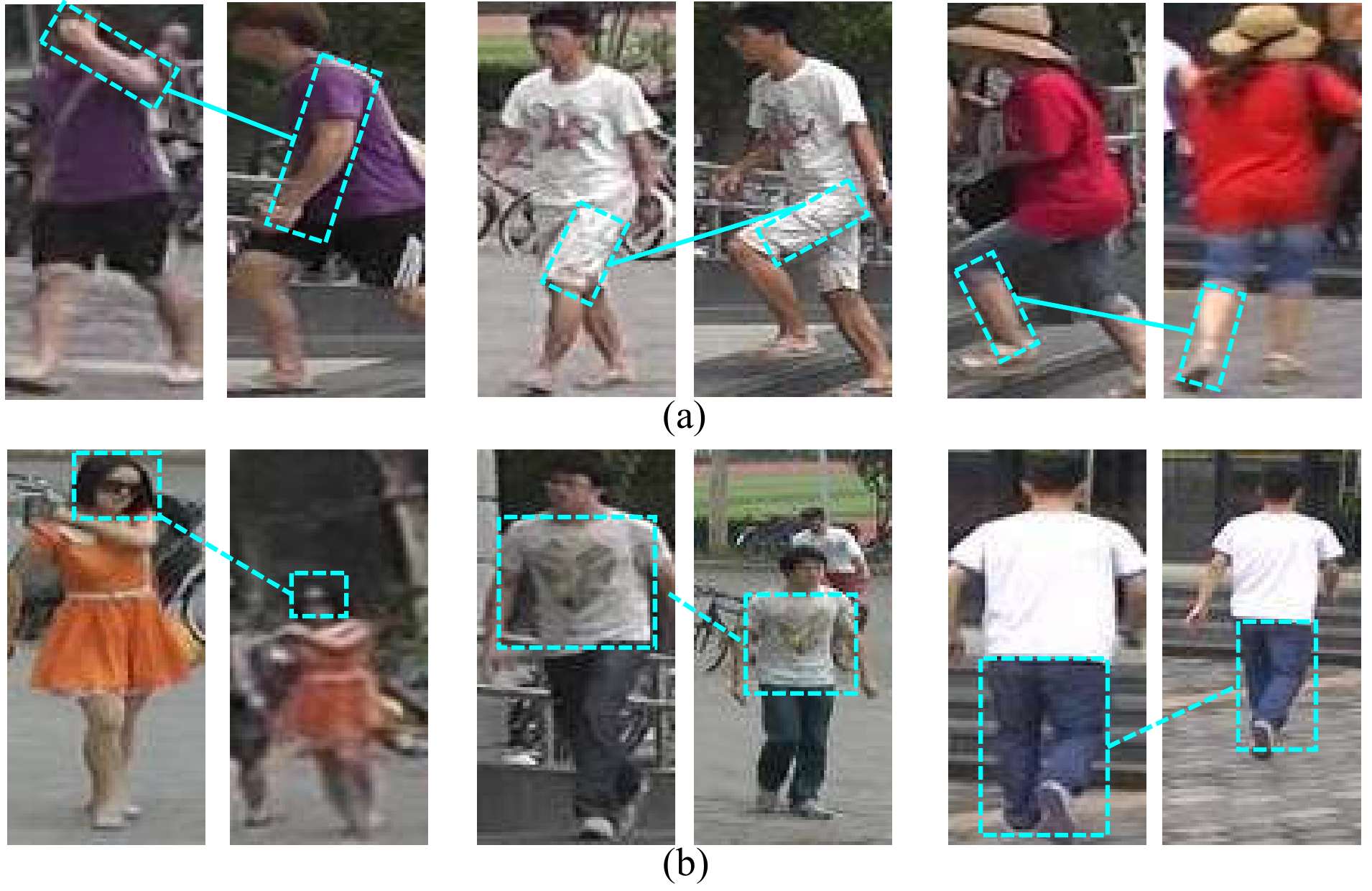}}
     \caption{Example image pairs that illustrate the challenges faced by ReID when attempting to extract semantically aligned part-level representations. (a) Both the appearance and location of body parts change due to variations in viewpoint and human pose. (b) The position of body parts is affected by errors in pedestrian detection. Green rectangles indicate the same body part.}
    \label{examples}
    \end{figure}

    The key to robust ReID lies in high-quality pedestrian representations. Recently, part-level representations \cite{ECCVsuh2018part, CVPRzhang2019densely, ECCVsun2018beyond,CVPRli2018harmonious, TIPyao2019deep, TIPPart-based2017,CVPRzheng2019pyramidal, CVPRzhao2017spindle, zhou2019robust} have become more popular, as they include fine-grained features and can effectively alleviate the overfitting risk associated with deep models. However, as illustrated in Fig. \ref{examples}, part misalignment frequently occurs in pedestrian images; this is caused by variations in viewpoint and human pose \cite{CVPRsaquib2018pose, ICCVsu2017pose}, as well as errors in pedestrian detection.

    Many previous works have attempted to solve the semantic misalignment problem in the context of body parts \cite{TIPyao2019deep,TIPPart-based2017,CVPRzheng2019pyramidal, CVPRyang2019patch, CVPRli2018harmonious, CVPRzhao2017spindle, ECCVsuh2018part, ECCVsun2018beyond}. One intuitive strategy involves detecting body parts during both training and testing before feature extraction \cite{CVPRli2018harmonious, TIPyao2019deep, CVPRli2017learning, ICCVsu2017pose}. However, there are two key downsides to this approach: first, part detection introduces an additional computational cost; second, part detection is inherently challenging, particularly for low-quality images that suffer from severe image blur or occlusion. Several recent works \cite{CVPRzhang2019densely, factor2018} have successfully bypassed part detection in the inference phase. These works mainly adopt the \emph{teacher-student} training strategy, in which the teacher model still relies on part location priors. During the inference process, only the student model is required for ReID. There are, however, several drawbacks of this method: for example, the architecture of the model during training is complex \cite{CVPRzhang2019densely}, and accurate location of body parts for each training image is still required \cite{CVPRzhang2019densely}.

    Given the above, it is therefore reasonable to ask whether part detection can be bypassed during both the training and testing phases. Our key observation here is that statistics in a batch of images are stable. The quantitative analysis for this criterion is provided in the supplementary material. Therefore, we propose a simple framework, named Batch Coherence-Driven Network (BCD-Net), which learns semantically aligned part-level representations by exploiting batch coherence from both channel and spatial perspectives. More specifically, we build $K$ sub-networks on top of a deep backbone model \cite{he2016deep, ECCVsun2018beyond}; each of these sub-networks extracts features for one specific body part with the help of two complementary components, namely a batch coherence-guided channel attention (BCCA) module and a pair of spatial regularization terms.

    Each sub-network includes one BCCA module, which highlights the relevant channels for one body part from the output of the deep backbone model. It has been shown in previous works that one channel in the top layers of Convolutional Neural Networks (CNNs) describes a particular visual pattern \cite{TIPyao2019deep, CVPRli2018diversity, wang2019cdpm}, i.e., different body parts in ReID; in other words, there is correspondence between body parts and channels. However, part-channel correspondence is not provided. Accordingly, we here propose a statistically robust method to estimate the part-channel associations from each batch of training images. Based on this part-channel correspondence, we impose a supervision signal that guides the optimization of BCCA parameters. By using this method, BCCA is able to highlight channels that are relevant to one specific part and suppress the irrelevant ones. The supervision is imposed on the averaged output by BCCA in each batch, which allows the output of BCCA to be adjustable for each individual image. Finally, because BCCAs in different sub-networks receive different supervision signals, these sub-networks are able to extract diverse and semantically aligned part-level features.

    Furthermore, we introduce a pair of spatial regularization terms that are complementary to BCCA. First, we observe that although the location of a single body part in each image varies, its mean position in a single batch of training images remains stable; this means that the average location of one body part is coherent across batches. Accordingly, we impose one part-level regularization on the feature maps for each respective sub-network. More specifically, for each sub-network, we conduct average pooling for the feature maps of all training images in one batch, then constrain the high-response areas to be consistent with the part's default location in well-aligned images. Moreover, by encouraging the $K$ sub-networks to respond strongly on different locations, these sub-networks are regularized to enable the learning of diverse part-level features. Since the part-level regularization is applied independently to each body part, information may be lost between adjacent parts. We therefore propose the holistic-level regularization term to handle this problem. This term requires that the aggregate of the responses of these $K$ sub-networks cover the full human body, with the results that all locations on the human body are encouraged to respond. Consequently, the two regularization terms help BCD-Net to learn diverse and complementary part-level features.


    During the testing process, all three constraints are removed. BCD-Net can automatically extract part-aware representations, since the parameters of each of the $K$ sub-networks have been optimized to be sensitive to the features of one specific part.

    From the methodological point of view, the contributions of this work can be summarized as follows. First, we introduce a novel model that extracts semantically aligned part features for ReID. Different from existing part-based ReID methods that typically employ two separate steps, namely, body part detection and part-level feature extraction, our model completely bypasses the part detection step. Second, we introduce the batch coherence criterion, according to which we design constraints from both channel and spatial perspectives. To the best of our knowledge, this is the first attempt that employs the batch coherence criterion for part-aware ReID.

    Extensive experiments justify the effectiveness of each key component in BCD-Net. More impressively, BCD-Net consistently achieves state-of-the-art performance on four large-scale ReID benchmarks: Market-1501 \cite{zheng2015scalable}, DukeMTMC-reID \cite{zheng2017unlabeled}, CUHK03 \cite{li2014deepreid}, and MSMT17 \cite{wei2018person}.

    The remainder of this paper is organized as follows. Related works on part-based ReID are briefly reviewed in Section II. The BCD-Net model structure and training scheme are described in Section III. Detailed experiments and their analysis are presented in Section IV. Finally, we conclude this paper in Section V.

\section{Related Work}

\subsection{Part-based ReID Models}
    In recent years, CNNs have been widely applied to ReID \cite{ECCVsun2018beyond, CVPRli2018harmonious, CVPRli2018diversity, CVPRzhang2019densely, CVPRyang2019patch, TIPyao2019deep, TIPPart-based2017, ICCVChen2019ABD-Net, CVPRyang2019towards, song2020unsupervised, yang2020part, zhong2019invariance,mpn}. In particular, part-based deep representations \cite{TIPyao2019deep, CVPRzhao2017spindle,CVPRzhang2019densely,CVPRzheng2019pyramidal, CVPRyang2019patch, CVPRli2018harmonious, ECCVsun2018beyond,mpn} have been proven to be effective for ReID, as they alleviate the overfitting risk associated with deep models and their features contain fine-grained information. Early methods typically extracted part features from fixed spatial locations \cite{fu2019horizontal, wang2018learning}, meaning that their performance became sensitive to the variations of part locations. Accordingly, many subsequent works \cite{CVPRyang2019patch, CVPRli2018harmonious, CVPRzhao2017spindle, CVPRli2017learning} aimed to extract semantically aligned part-level representations. These methods can be divided into two categories according to whether or not part detection is required during training and testing.

\subsubsection{Part Detection During Both Training and Testing}
    Methods in this category \cite{TIPyao2019deep,TIPPart-based2017,CVPRzhao2017spindle, CVPRyang2019patch, CVPRli2018harmonious, CVPRli2017learning, ECCVsuh2018part, ECCVsun2018beyond, wang2019cdpm} perform body part detection and part-level feature extraction sequentially during both the training and testing phases. Part detection provides the spatial location of body parts, usually in the form of bounding boxes \cite{CVPRzhao2017spindle, TIPyao2019deep, ICCVsu2017pose, wang2019cdpm}, from which the part-level features are extracted. There are two popular strategies commonly adopted for part detection. The first of these utilizes outside tools, e.g., pose estimation models \cite{ICCVsu2017pose, CVPRzhao2017spindle, CVPRxu2018attention}, human parsing algorithms \cite{CVPRkalayeh2018human} or human segmentation models \cite{wang2019cdpm}, to detect body parts. For example, Wang \emph{et al.} \cite{wang2019cdpm} adopted a human segmentation model to produce the upper and lower boundaries of a pedestrian. The body parts are obtained by uniformly dividing the pedestrian between the boundaries in the training stage. However, there are two downsides of this approach: first, it requires additional computational cost; second, the ReID performance can be negatively impacted by the low reliability of outside tools. The second strategy adopts an attention mechanism to infer the location of body parts according to the feature maps of ReID models \cite{CVPRli2018harmonious, ICCVzhao2017deeply, CVPRli2017learning, liu2017hydraplus}. While attention-based methods tend to be more efficient than outside tool-based methods, they still face challenges due to interference in low-quality images (e.g., severe image blur and background clutter). In addition, one existing work \cite{TIPyao2019deep} clustered channels in feature maps based on the locations of their maximum responses, such that the average response in each cluster of channels indicates the position of a particular part. However, this strategy is sensitive to the part missing problem that arises due to pedestrian detection errors.

\subsubsection{Part Detection During Training Only}
    To address the aforementioned problems, some methods have been recently proposed that bypass part detection during testing \cite{CVPRzhang2019densely, factor2018}. For example, Zhang \emph{et al.} \cite{CVPRzhang2019densely} leveraged a 3D person model to construct a set of semantically aligned part images for each training image. These part images enable a teacher model to learn semantically aligned part features. Subsequently, the teacher model transfers the body part concepts to a student model via alignment in their feature spaces. In the testing phase, the student model can then independently extract part-aware features. However, despite the convenience this affords during testing, this approach still relies on the complex prior information of body part locations for each training image.

    For its part, the BCD-Net method proposed in this paper bypasses body part detection during both the training and testing phases, but can still extract semantically aligned part features. Therefore, it is both robust and easy-to-use for real-world applications.

\subsection{Methods Based on Batch-level Information}
    Over the past few years, the use of batch-level information has been successfully explored in the fields of deep learning and computer vision \cite{ioffe2015batch, Singh2019EvalNorm, Sergey2017batch, chang2019domain, li2017revisiting}. For example, batch normalization \cite{ioffe2015batch} (BN) has significantly accelerated the training of deep models by normalizing features using the mean and variance computed from the entire mini-batch. Ioffe \emph{et al.} \cite{Sergey2017batch} improved BN by constraining the moments of a mini-batch to a specific range, thereby reducing the variation of statistics during training.

    Inspired by BN, several works have proposed adopting batch-level information for domain adaptation \cite{chang2019domain, li2017revisiting}. For example, Li \emph{et al.} \cite{li2017revisiting} argued that domain knowledge can be reflected by the statistics in the BN layer. Accordingly, they performed domain adaptation by replacing the statistics of the source domain with those of the target domain, which promotes the model's generalization ability.

    For our part, we utilize batch-level information in another application, namely part-aware ReID. In brief, we make use of the batch-level information to build constraints from both channel and spatial perspectives. These constraints drive ReID models to extract semantically aligned part-level features. We have accordingly named the proposed method Batch Coherence-Driven Network (BCD-Net).

\section{Batch Coherence-Driven Network}
    In this section, we first introduce the motivation and overall network architecture of BCD-Net, then describe each of its key components in detail: namely, sub-networks for feature extraction, part-relevant channel identification via BCCA, and a pair of spatial regularization terms.

\subsection{Overview}
\begin{figure}[t]
    \centerline{\includegraphics[width=0.50\textwidth]{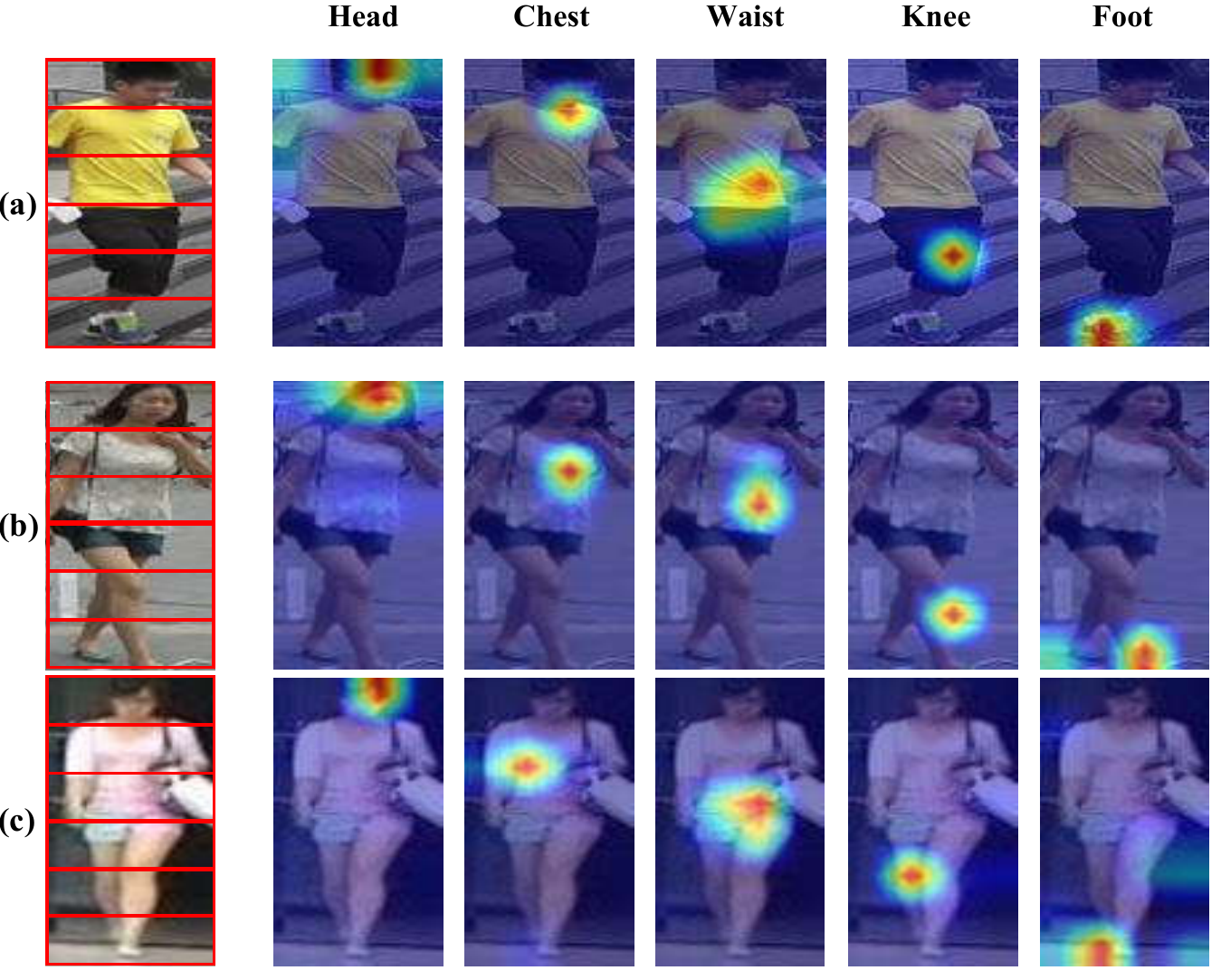}}
     \caption{Visualization results of responses on five representative channels in the feature maps output by a ResNet-50 backbone model. Each row presents responses for the same image. It can be seen that there exists correspondence between the channels and body parts; in other words, each channel in this figure describes one body part.}
    \label{channelvis}
    \end{figure}

    Existing works have demonstrated that specific channels in the top layers of CNNs describe specific visual patterns \cite{zheng2017learning, simon2015neural, zhang2016picking, TIPyao2019deep, wang2020multistage} (i.e., different body parts in ReID). As explained in Fig. \ref{channelvis}, each column represents responses on the same channel for different images. Here, we present five representative channels, each of which respectively corresponds to the head, chest, waist, knee, and foot of one pedestrian. It is evident that correspondence exists between channels and body parts; this motivates us to extract part-level features based on part-relevant channels. In this paper, we define body parts according to the method outlined in \cite{ECCVsun2018beyond, wang2019cdpm}. In brief, we divide one pedestrian uniformly into $K$ parts in the vertical direction to represent their position, as illustrated in Fig. \ref{channelvis}.


\begin{figure*}[t]
    \centerline{\includegraphics[width=1.0\textwidth]{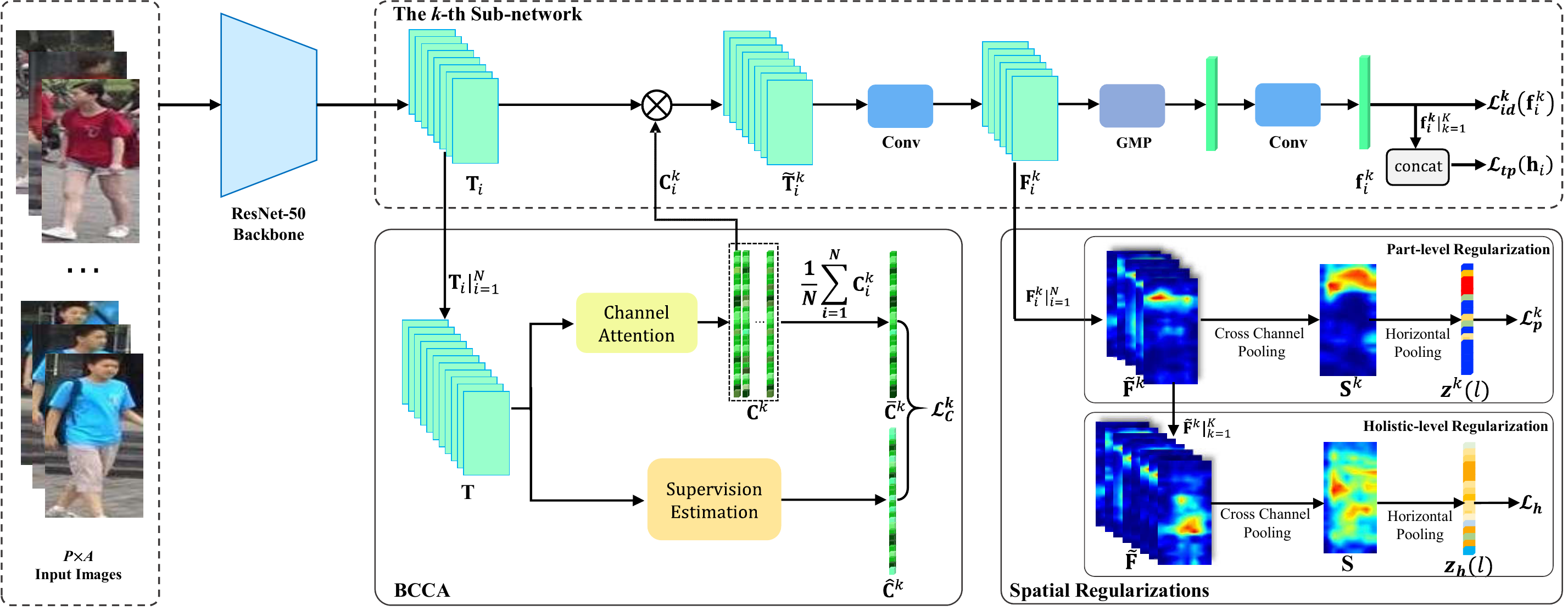}}
     \caption{Architecture of BCD-Net during the training stage. BCD-Net builds $K$ sub-networks on the ResNet-50 backbone model for $K$ body parts. In the
     interests of simplicity, only one sub-network is illustrated in this figure; the other sub-networks share the same structure. BCD-Net is free from body part detection during both training and testing, but still extracts semantically aligned part-level features. Therefore, it is very efficient and easy to implement. Semantic consistency of part features is ensured by two complementary components, i.e., BCCA and a pair of spatial regularization terms. We concatenate the $K$ part features as the pedestrian representation.}
    \label{framework}
    \end{figure*}

    We build $K$ sub-networks on the top of a deep backbone model \cite{he2016deep}, each of which extracts features for a specific body part with the help of two complementary components, i.e., BCCA and a pair of spatial regularization terms. The architecture of BCD-Net is illustrated in Fig. \ref{framework}. Following the method outlined in \cite{ECCVsun2018beyond}, we use ResNet-50 \cite{he2016deep} as the backbone model and remove its last spatial down-sampling operation to increase the size of the output feature maps. In the interests of simplicity, the output feature maps for the $i$-th image in the batch are denoted as $\mathbf{T}_i\in {\mathbb{R}}^{C\times H \times W}$ in the following; here, these three numbers denote the channel, height, and width dimensions respectively. The aggregation of feature maps of $N$ images in the batch is denoted as a four-dimensional tensor $\mathbf{T} \in {\mathbb{R}}^{N \times C\times H \times W}$. Next, we attach $K$ sub-networks to the backbone model in order to perform part-level feature extraction. The semantic consistency of the part features is ensured by the two complementary components, i.e., BCCA and a pair of spatial regularization terms.

\subsection{Sub-networks for Feature Extraction}
    As illustrated in Fig. \ref{framework}, the structure of each sub-network, which is constructed in a similar way to that proposed in two recent works \cite{CVPRzhang2019densely, wang2019cdpm}, comprises the following: one $1 \times 1$ convolutional (Conv) layer, one global maximum pooling (GMP) layer, a second $1 \times 1$ Conv layer, and one fully connected (FC) layer for person classification. Moreover, both Conv layers incorporate one BN layer \cite{ioffe2015batch} and one ReLU layer \cite{nair2010rectified}. The dimension of both Conv layers is set to 512. Each of the $K$ sub-networks is optimized by the cross-entropy loss attached to the FC layer. The loss function for the $k$-th sub-network is formulated as follows:
    \begin{equation}
    {\mathcal{L}}_{id}^{k} = - \frac{1}{N}\sum_{i=1}^N{\log{\frac{e^{{\mathbf{w}_{y_{i}}^{k}}^{\mathsf{T}} {\bf{f}}^{k}_{i}}}{\sum_{j=1}^{J}{e^{{\mathbf{w}_{j}^{k}}^{\mathsf{T}}{\bf{f}}^{k}_{i}
    }}}}},
    \end{equation}
    where $\mathbf{w}_{j}^{k}$ is the weight vector for class $j$, \textit{N} is the batchsize during training, and \textit{J} is the number of classes in the training set. Moreover, $y_{i}$ and ${\bf{f}}_{i}^{k}$ represent the label and the $k$-th part feature for the $i$-th image in a batch, respectively. The output of the second $1 \times 1$ Conv layer is chosen as the part feature. The bias term is omitted in the interests of simplicity.

    We concatenate the $K$ part features to create the holistic representation $\mathbf{h}_{i}$ for the $i$-th image, as follows:
    \begin{equation}
    \mathbf{h}_{i} = [{\mathbf{f}_i^{1}}^{\mathsf{T}}~{\mathbf{f}_i^{2}}^{\mathsf{T}}~\dots~{\mathbf{f}_i^K}^{\mathsf{T}}]^{\mathsf{T}}.
    \label{concat}
    \end{equation}

    $\mathbf{h}_{i}$ is further optimized by the triplet loss function \cite{schroff2015facenet} with a batch-hard triplet sampling policy \cite{hermans2017defense}. In order to ensure that sufficient triplets are sampled during training, we randomly sample $A$ images in each of $P$ random identities to compose a batch; therefore, the batch size $N$ is equal to $P \times A$. The triplet loss is formulated as follows: ${\mathcal{L}}_{tp}=$
    \begin{equation}
    \frac{1}{N_{tp}}\sum_{i=1}^P\sum_{a=1}^A
    [\max_{p=1...A} \mathcal{D} (\textbf{h}_i^{a}, \textbf{h}_i^{p}) - \min_{\substack{n=1...A \\ j=1...P \\ j \neq
    i}}\mathcal{D} (\textbf{h}_i^{a}, \textbf{h}_j^{n})  + \alpha ]_+,
    \label{tripletloss}
    \end{equation}
    where $\alpha$ is the margin of the triplet constraint, while $N_{tp}$ denotes the number of triplets that violate the triplet constraint in a given batch. $[\cdot]_+ = \max(\cdot,0)$ represents the hinge loss. $\textbf{h}_i^{a}$, $\textbf{h}_i^{p}$ and $\textbf{h}_j^{n}$ are holistic representations of the anchor, positive, and negative images in a triplet, respectively. Finally, $\mathcal{D} (\textbf{h}_i, \textbf{h}_j)$ denotes the cosine distance between the two feature vectors $\textbf{h}_i$ and $\textbf{h}_j$.

\subsection{Batch Coherence-guided Channel Attention}
    The $K$ sub-networks share identical input: i.e., $\mathbf{T}_{i}$. Naive training of the above model results in all sub-networks extracting similar holistic features, as illustrated in Fig. \ref{visualization1}(a). Existing works solve this problem by providing the sub-networks with the location of body parts for each image, either during training only \cite{CVPRzhang2019densely, factor2018} or during both training and testing \cite{TIPyao2019deep, CVPRzhao2017spindle, CVPRli2018harmonious, CVPRli2017learning, ECCVsuh2018part}. However, body part detection is challenging, especially for low-quality images.

\begin{figure}[t]
    \centerline{\includegraphics[width=0.5\textwidth]{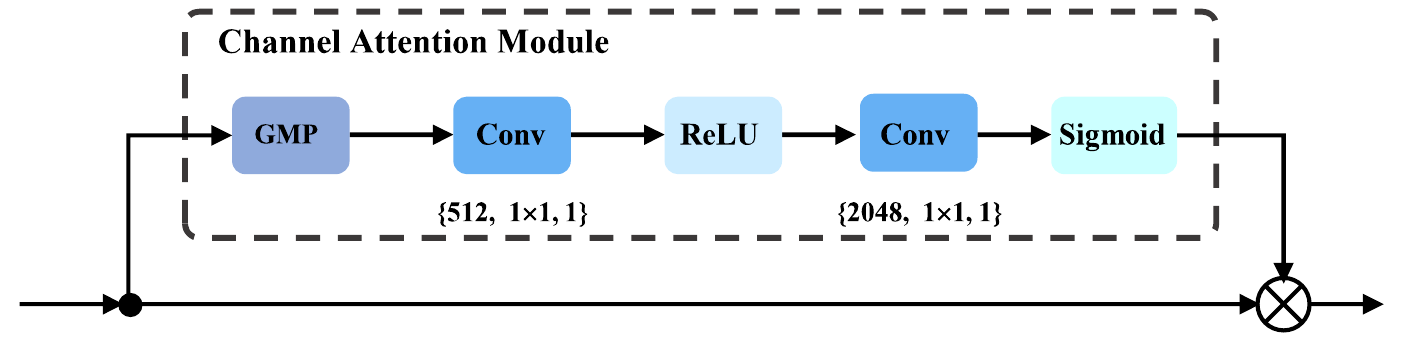}}
     \caption{Structure of the adopted channel attention module. The items in each bracket denote the number of filters, kernel size, and stride respectively. By default, each Conv layer is followed by a BN layer.}
     \label{ca}
    \end{figure}

    We solve this problem through the use of BCCA, which highlights the relevant channels for each individual body part. As shown in Fig. \ref{framework}, BCCA is embedded into each sub-network. Similar to existing channel attention (CA) models \cite{CVPRli2018harmonious, secvpr2018}, BCCA is realized by a GMP layer and two successive $1 \times 1$ Conv layers. The configuration of BCCA is illustrated in more detail in Fig. \ref{ca}. Moreover, ${\bf{C}}_i^k$ in Fig. \ref{framework} denotes the predicted channel weights for the $i$-th image in a batch and is utilized to recalibrate channels in $\mathbf{T}_{i}$ via channel-wise multiplication, as follows:
    \begin{equation}
    \widetilde{\mathbf{T}}^{k}_{i} = \mathbf{C}^{k}_{i} \otimes \mathbf{T}_{i},
    \label{multiplication}
    \end{equation}
    where $\otimes$ denotes the channel-wise multiplication operation.

    However, without explicit guidance, BCCA is unable to automatically identify part-relevant channels. Therefore, we introduce a specific supervision signal $\hat{\mathbf{C}}^{k}$ for the BCCA module in the $k$-th sub-network. The elements in $\hat{\mathbf{C}}^{k}$ indicate the relevance of each channel to the $k$-th part. We estimate $\hat{\mathbf{C}}^{k}$ from the images in each training batch according to the following four steps.

    First, we reshape $\mathbf{T} \in {\mathbb{R}}^{N \times C\times H\times W}$ into a set of $C$ feature maps ${\mathbf{M}}_{c} \in {\mathbb{R}}^{N\times H\times W}$. ${\mathbf{M}}_{c}$ is constructed by stacking the $c$-th channel in $\mathbf{T}_{i}$ ($1\leqslant i\leqslant N$) according to the channel dimension.

\begin{figure}[t]
    \centerline{\includegraphics[width=0.5\textwidth]{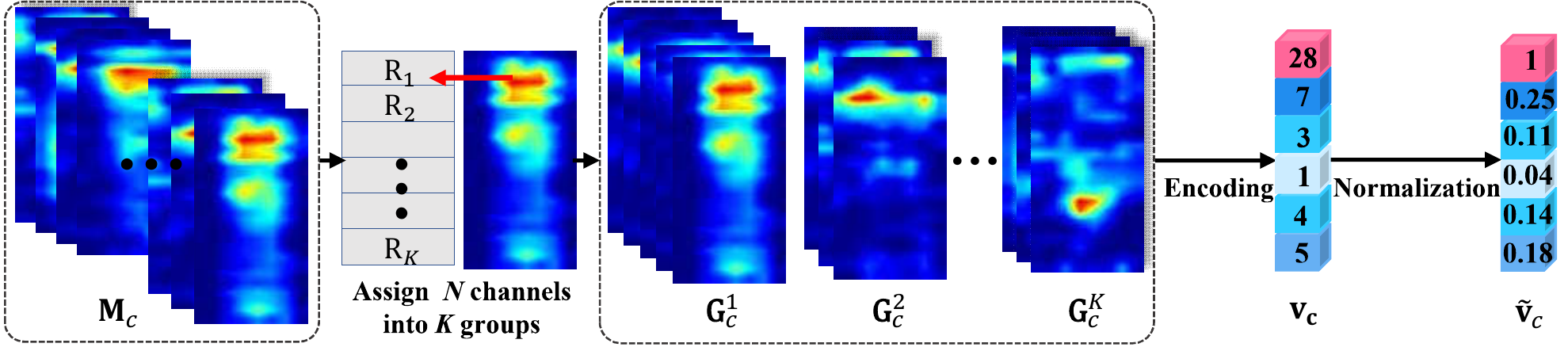}}
     \caption{The pipeline for estimating the relevance of the $c$-th channel in $\mathbf{T}_{i}$ to $K$ body parts. We first divide the $N$ channels in ${\mathbf{M}}_{c}$ into $K$ groups $\mathbf{\mathbf{G}}^{k}_{c}$ ($1\leqslant k\leqslant K$) based on the location of their maximum responses. Subsequently, $\mathbf{G}^{k}_{c}$ ($1\leqslant k\leqslant K$) are encoded into a vector $\mathbf{{v}}_{c}$, which is then normalized to produce $\mathbf{\widetilde{v}}_{c}$.}
     \label{supervision_group}
    \end{figure}

    Second, we estimate the relevance of each channel to each of the $K$ body parts. The extent of this relevance can be inferred from the frequency that the channel activates in the region of each respective part. Although the part location for each image is unknown, we can observe that the alignment errors in most images are small or moderate. Accordingly, we first roughly estimate the part location by uniformly dividing each channel in ${\mathbf{M}}_{c}$ into $K$ regions ${\rm {R}}_{k}$ ($1\leqslant k\leqslant K$), as illustrated in Fig. \ref{supervision_group}. Subsequently, the $N$ channels in ${\mathbf{M}}_{c}$ are divided into $K$ groups based on the location of their maximum responses. More specifically, we assign a channel to the $k$-th group $\mathbf{G}^{k}_{c}$ when the spatial location of its maximum response lies in ${\rm {R}}_{k}$. Therefore, we can encode ${\mathbf{M}}_{c}$ into $\mathbf{v}_{c}$ as follows:
    \begin{equation}
    \mathbf{v}_{c} = [{{v}_c^{1}}, {{v}_c^{2}}, \dots, {{v}_c^K}],
    \label{supervision}
    \end{equation}
    where ${v}_c^k$ denotes the number of channels in $\mathbf{G}^{k}_{c}$. Intuitively, $\mathbf{v}_{c}$ indicates the relevance of the $c$-th channel in $\mathbf{T}_{i}$ to each body part.

    Third, we normalize $\mathbf{v}_{c}$ to obtain $\mathbf{\widetilde{v}}_{c}$, as follows:
    \begin{equation}
    \mathbf{\widetilde{v}}_{c}=\left\{\begin{array}{ll}
    {[0, 0, ... , 0]} & {\text { if } {\ddot{v}}_c < \lfloor \beta \times N \rfloor} \\
    {[\frac{{v}_c^{1}}{{\ddot{v}}_c}, \frac{{v}_c^{2}}{{\ddot{v}}_c}, \dots, \frac{{v}_c^{K}}{{\ddot{v}}_c}]} & {\text { if } {\ddot{v}}_c \geqslant \lfloor \beta \times N \rfloor},\end{array}\right.
    \label{supervision2}
    \end{equation}
    where ${\ddot{v}}_c = max \{{{v}_c^{1}}, {{v}_c^{2}}, \dots, {{v}_c^K}\}$. $\beta$ is a pre-defined hyper-parameter and $\lfloor * \rfloor$ represents the rounding down operation on $*$. If the value of ${\ddot{v}}_c$ is smaller than $\lfloor \beta \times N \rfloor$, we regard the $c$-th channel as an irrelevant channel that will not be useful for describing a specific body part. When this occurs, we perform filtration via setting the $K$ elements in $\mathbf{\widetilde{v}}_{c}$ to 0; otherwise, we normalize $\mathbf{v}_{c}$ using ${\ddot{v}}_c$, as described in Eq. \ref{supervision2}.

    Finally, we generate the supervision signal $\hat{\mathbf{C}}^{k}$ for the output of BCCA. We concatenate $\mathbf{\widetilde{v}}_{c}$ ($1\leqslant c\leqslant C$) along the row dimension and obtain a matrix $\hat{\mathbf{C}} \in {\mathbb{R}}^{C \times K}$. The $k$-th column of $\hat{\mathbf{C}}$ is the supervision $\hat{\mathbf{C}}^{k}$ for BCCA that reflects the relevance of the $C$ channels to the $k$-th part.

    Rather than imposing the above supervision signal on each individual image, we instead propose a batch-level constraint that makes the output of BCCA adjustable for each image. More specifically, we average the output of BCCA for $N$ images as follows:
    \begin{equation}
    {\bar{\bf{C}}}^{k} = \frac{1}{N} \sum_{i=1}^{N} {\bf{C}}^{k}_{i}.
    \label{ca2}
    \end{equation}
    Next, we optimize the output of BCCA by minimizing the cosine distance between $\bar{\mathbf{C}}^{k}$ and $\hat{\mathbf{C}}^{k}$, as follows:
    \begin{equation}
    {\mathcal{L}}_{C}^{k} = \mathcal{D} (\bar{\mathbf{C}}^{k}, \hat{\mathbf{C}}^{k}).
    \label{bcca}
    \end{equation}

    Since we have imposed different supervision signals on the $K$ BCCA modules, the features extracted by $K$ sub-networks become part-specific and therefore diverse.

\subsection{Spatial Regularizations}
    In the next step, we propose a pair of spatial regularization terms that are complementary to BCCA. We can observe that, although the location of the same body part varies from image to image, its average location across all training images in a batch is stable. We therefore propose a pair of simple batch-level regularization terms, which guide the $K$ sub-networks to learn diverse and complementary part-level features while also bypassing part detection for each image.

\subsubsection{Part-level Regularization}
\begin{figure}[t]
    \centerline{\includegraphics[width=0.5\textwidth]{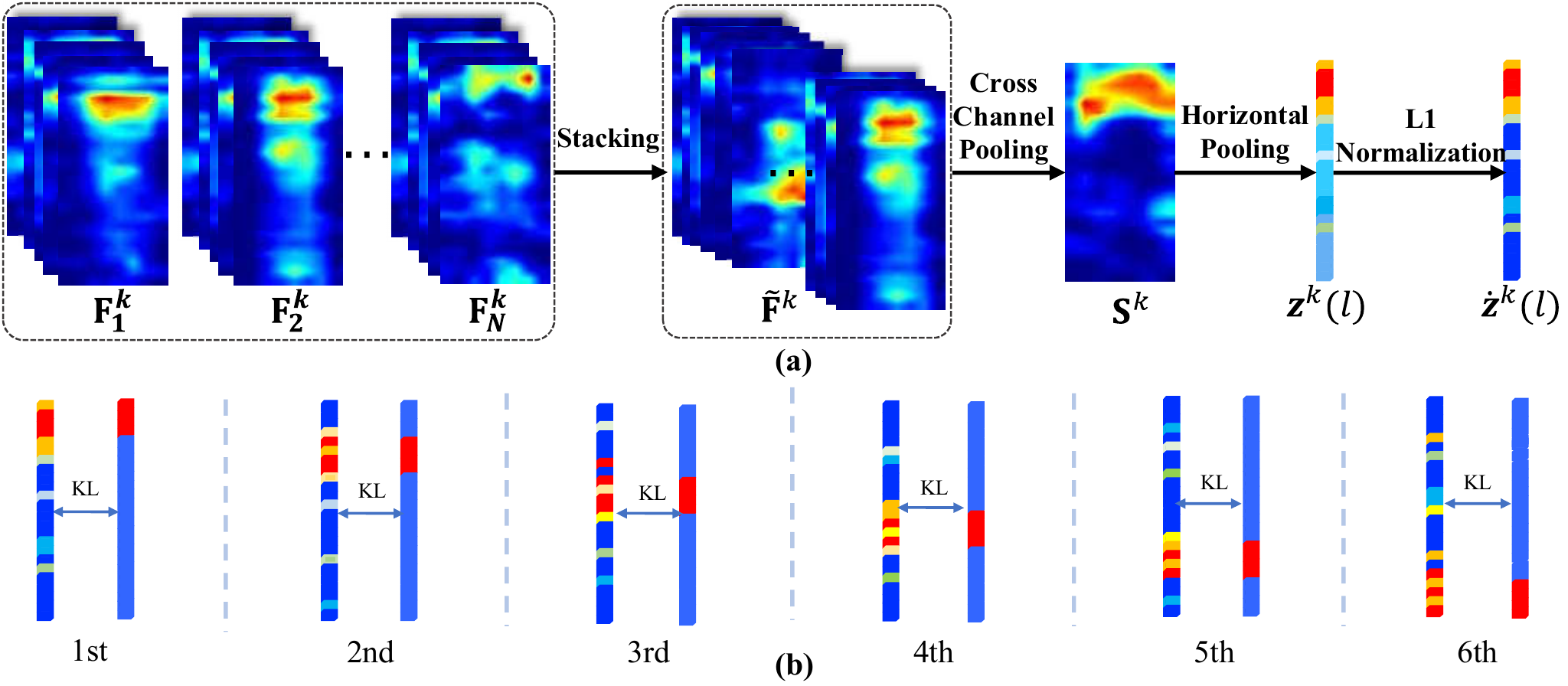}}
     \caption{Illustration of how the part-level regularization term is applied to body parts. (a) Flowchart of computing $\bold{\dot{z}}^k(l)$ for a batch of images. (b)  For each sub-network, KL-Divergence loss is employed to minimize the discrepancy between the predicted response $\bold{\dot{z}}^k(l)$ (on the left) and ground-truth response $\bold{\hat{z}}^{k}(l)$ (on the right). Red and blue denote stronger and weaker responses, respectively. This figure depicts the regularization terms when $K$ is set to 6. (Best viewed in color.)}
    \label{part_kl}
    \end{figure}

    This regularization is imposed on each sub-network. For the $k$-th sub-network, the output feature maps of its first Conv layer after BCCA can be denoted as $\mathbf{F}^k\in {\mathbb{R}}^{N\times C_{1} \times H\times W}$ for a batch; here, the four numbers indicate the batch, channel, height, and width dimensions, respectively. Moreover, the feature maps for the $i$-th training image are denoted as $\mathbf{F}^k_{i}\in {\mathbb{R}}^{C_{1} \times H\times W}$. We reshape $\mathbf{F}^k$ into $\mathbf{{\widetilde{F}}}^k\in {\mathbb{R}}^{U\times H\times W}$, where $U=N\times C_{1}$. As shown in Fig. \ref{part_kl}(a), we obtain a spatial attention map $\mathbf{S}^k\in {\mathbb{R}}^{ H\times W}$ by performing average pooling on $\mathbf{{\widetilde{F}}}^k$ along the channel dimension, and subsequently obtain the vector $\bold{z}^k(l)\in {\mathbb{R}}^{H}$ by averaging the values for each row of $\mathbf{S}^k$.
    More formally:
    \begin{equation}
    \bold{z}^k(l)=\frac{1}{U \times W} \sum_{i=1}^{U} \sum_{j=1}^{W} \mathbf{{\widetilde{F}}}^k_{(i,l,j)}.
    \end{equation}
    Here, $\mathbf{{\widetilde{F}}}^k_{(i,l,j)}$ represents the element that lies in the $i$-th channel, $l$-th row, and $j$-th column of $\mathbf{{\widetilde{F}}}^k$. Accordingly, each element in $\bold{z}^k(l)$ represents the average response of images in a batch at a specific position along the height dimension. We then normalize $\bold{z}^k(l)$ using its L1 norm, as follows:
    \begin{equation}
    \bold{\dot{z}}^k(l) = \frac{\bold{z}^k(l)}{\|\bold{z}^k(l)\|_1}.
    \end{equation}

    We anticipate that elements with a high response in $\bold{\dot{z}}^k(l)$ will be consistent with the default location of the part in question in well-aligned images. Therefore, the ground-truth response vector $\bold{\hat{z}}^{k}(l)$ can be formulated as follows:
    \begin{equation}
    \bold{\hat{z}}^{k}(l)=\left\{\begin{array}{ll}
    {\gamma} & {\text {if } \frac{ H
    \times (k-1)}{K}+1\leqslant l\leqslant\frac{ H \times k}{K}} \\
    {\frac{K - \gamma \times H} {H \times (K-1)}} & {\text {otherwise, }}\end{array}\right.
    \label{part_kl_label}
    \end{equation}
    where $\gamma$ is a hyper-parameter used to control the amplitude of the part features in the height dimension. The shape of $\bold{\hat{z}}^{k}(l)$ is illustrated in Fig. \ref{part_kl}(b). We assign high values (i.e., $\gamma$) to the $\frac{H}{K}$ elements that are relevant to the $k$-th part in $\bold{\hat{z}}^{k}(l)$. We also assign small values to the other elements in $\bold{\hat{z}}^{k}(l)$ to account for the variation in body part locations.

    As shown in Fig. \ref{part_kl}(b), $\bold{\hat{z}}^{k}(l)$ is different for each of the $K$ sub-networks to ensure diversity among the $K$ part-level features. We use KL-Divergence loss to minimize the discrepancy between $\bold{\dot{z}}^k(l)$ and $\bold{\hat{z}}^{k}(l)$ for each sub-network:
    \begin{equation}
    \mathcal{L}^k_{p} = \sum\limits_{l = 1}^H {\bold{\hat{z}}^k(l)}\log (\frac{\bold{\hat{z}}^{k}(l)}{\bold{\dot{z}}^k(l)}).
    \label{part_spatial}
    \end{equation}

\subsubsection{Holistic-level Regularization}
    Although the above part-level regularization terms drive the sub-networks to learn part-specific features, they ignore the distribution of responses throughout the whole human body. Accordingly, we propose a holistic-level regularization term that guides the $K$ sub-networks to extract features from all locations across the human body.

    As illustrated in Fig. \ref{framework}, we concatenate $\mathbf{{\widetilde{F}}}^{k}$ ($1\leqslant k\leqslant K$) along the channel dimension and obtain $\mathbf{\widetilde{F}}\in {\mathbb{R}}^{(K\times U)\times H\times W}$. Next, we conduct cross-channel average pooling on $\mathbf{\widetilde{F}}$ and obtain another holistic attention map $\mathbf{S}\in {\mathbb{R}}^{H\times W}$. Finally, we generate a vector $\bold{z}_{h}(l)\in {\mathbb{R}}^{H}$ by averaging the values for each row in $\mathbf{S}$. $\bold{z}_{h}(l)$ indicates the average response of all $K$ sub-networks on the human body. More formally:
    \begin{equation}
    \bold{z}_{h}(l)=\frac{1}{K \times U \times W} \sum_{k=1}^{K} \sum_{i=1}^{U} \sum_{j=1}^{W} \mathbf{{\widetilde{F}}}^k_{(i,l,j)}.
    \end{equation}
    We then conduct L1 normalization for $\bold{z}_{h}(l)$, as follows:
    \begin{equation}
    \bold{\dot{z}}_{h}(l) = \frac{\bold{z}_{h}(l)}{\|\bold{z}_{h}(l)\|_1}.
    \end{equation}

    Since we expect responses from all locations on the human body, the ground-truth response vector $\bold{\hat{z}}_{h}(l)$ can be defined as follows:
    \begin{equation}
    \bold{\hat{z}}_{h}(l)=\frac{1}{H}, \quad \text 1\leqslant l\leqslant H.
    \end{equation}

    Similar to the part-level regularization approach, we minimize the KL-Divergence between $\bold{\dot{z}}_{h}(l)$ and $\bold{\hat{z}}_{h}(l)$ as follows:
    \begin{equation}
    \mathcal{L}_{h}  =  \sum\limits_{l = 1}^H {\bold{\hat{z}}_{h}(l)}\log (\frac{{\bold{\hat{z}}_{h}(l)}}{\bold{\dot{z}}_{h}(l)}).
    \label{holistic_spatial}
    \end{equation}

\subsection{Person ReID by BCD-Net}
    The overall objective function of BCD-Net during training can be written as follows:
    \begin{equation}
    \mathcal{L} = \sum_{k = 1}^{K}\mathcal{L}_{id}^{k}+\mathcal{L}_{tp} + {\lambda}_{1} \sum_{k = 1}^{K}\mathcal{L}_{C}^k + {\lambda}_{2} (\mathcal{L}_{h} + \sum_{k = 1}^{K}\mathcal{L}_{p}^k),
    \end{equation}
    where ${\lambda}_{1}$ and ${\lambda}_{2}$ are the weights of the loss functions. For the sake of simplicity, they are consistently set to 1.

    During the testing stage, the representation of one image
    is obtained by concatenating the $K$ part-level features, as described in Eq. \ref{concat}. The cosine metric is adopted consistently in order to measure the similarity between the representations of two images ${\mathbf{h}_{1}}$ and ${\mathbf{h}_{2}}$:
    \begin{equation}
    \rho = \frac{{\mathbf{h}_{1}}^{\mathsf{T}}{\mathbf{h}_{2}}}{\left\|{\mathbf{h}_{1}}\right\|\left\|{\mathbf{h}_{2}}\right\|},
    \label{simi}
    \end{equation}
    {\color{blue}} where $\left\|*\right\|$ indicates the L2 norm of *.

\section{Experiments}
To demonstrate the effectiveness of BCD-Net, we conduct extensive experiments on four large-scale ReID benchmarks: namely, Market-1501 \cite{zheng2015scalable}, DukeMTMC-reID \cite{zheng2017unlabeled}, CUHK03 \cite{li2014deepreid}, and MSMT17 \cite{wei2018person}. The official evaluation protocol for each database is followed for all experiments. We adopt the widely used Rank-1 accuracy and mean Average Precision (mAP) as metrics for evaluation.

Market-1501 \cite{zheng2015scalable} comprises 32,668 pedestrian images captured by six cameras for 1,501 identities. Pedestrians were detected using the Deformable Part Model (DPM) \cite{felzenszwalb2009object}. This dataset is divided into a training set and a testing set: the former contains 12,936 images of 751 identities, while the latter is composed of a gallery set and a query set containing 19,732 and 3,368 images respectively from the remaining 750 identities.

DukeMTMC-reID \cite{zheng2017unlabeled} contains 36,441 images of 1,404 identities captured using eight high-resolution cameras. A total of 16,522 images of 702 identities are reserved as the training set, while images of the other 702 identities are used for testing. The testing set is further split into a gallery set containing 17,661 images and a query set including the other 2,268 images.

CUHK03 \cite{li2014deepreid} includes 14,097 pedestrian images of 1,467 identities. Images in this dataset were captured by two disjoint cameras. This dataset provides two types of bounding boxes: namely, human-annotated ones and automatically detected ones using DPM \cite{felzenszwalb2009object}. Both types of bounding boxes are used for the evaluation of our method. The new training/testing protocol detailed in \cite{CVPRzhong2017re} is adopted. In line with this protocol, images of 767 identities make up the training set, while images of the other 700 identities are used for testing.

MSMT17 \cite{wei2018person} consists of 126,441 pedestrian images of 4,101 identities in total. A camera network consisting of three outdoor cameras and twelve indoor cameras was used to construct this dataset. MSMT17 is divided into a training set containing 32,621 images of 1,041 identities, and a testing set comprising 93,820 images of 3,060 identities. The testing set is further split into a gallery set of 82,161 images and a query set of 11,659 images.

\subsection{Implementation Details}
\begin{figure}[t]
    \centerline{\includegraphics[width=0.495\textwidth]{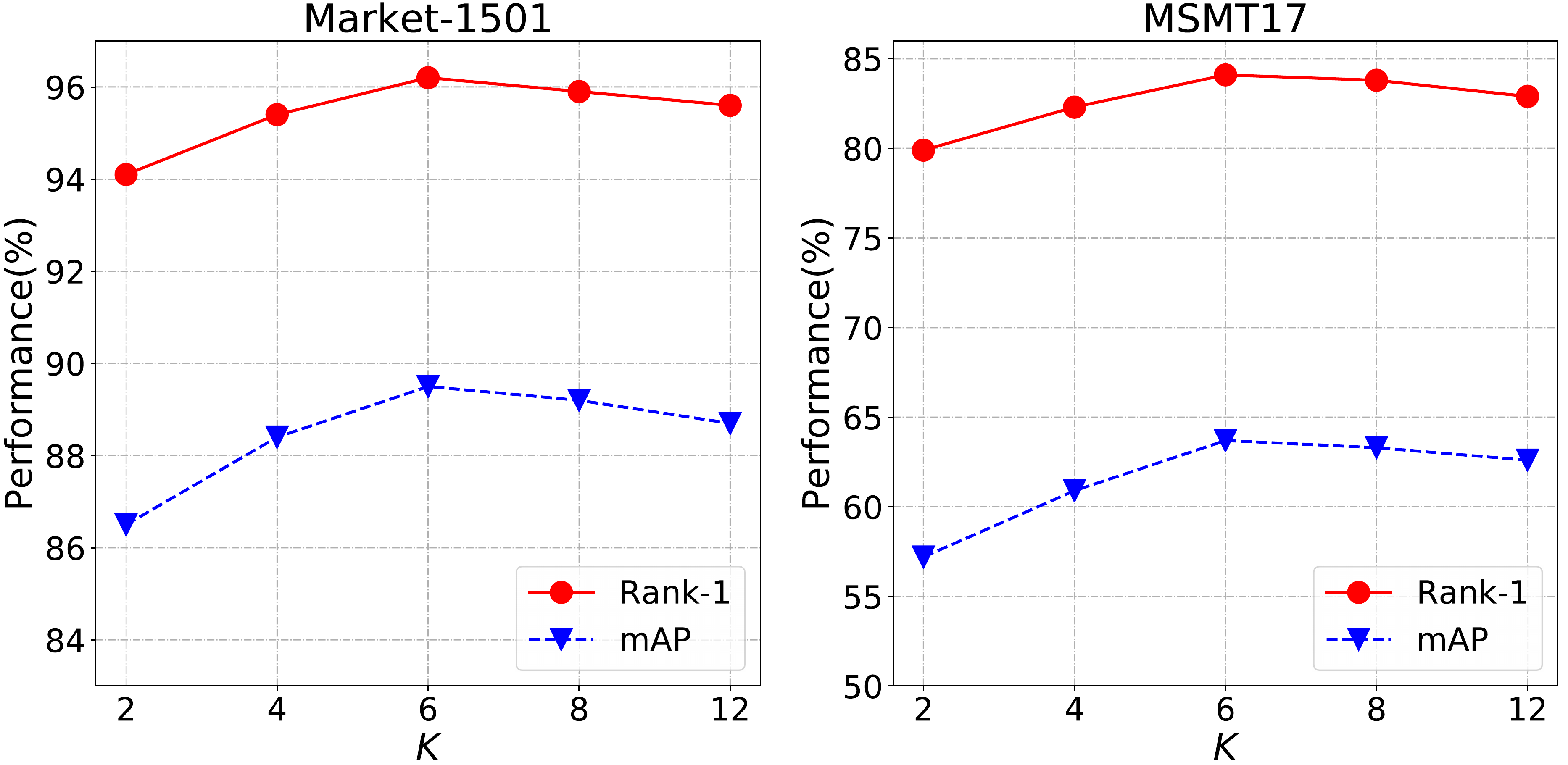}}
     \caption{Evaluation on the value of hyper-parameter $K$ for the performance of BCD-Net.}
    \label{K}
\end{figure}

The training sets of all benchmarks are augmented by means of offline translation \cite{li2014deepreid}, online horizontal flipping, and random erasing \cite{zhong2017random}. The use of offline translation enlarges each training set by a factor of 5. The ratio of random erasing is empirically set to 0.5. All pedestrian images are resized to $384 \times 128$ pixels. We set $P$ to 6 and $A$ to 8 in order to construct a batch (whose size is therefore 48). The hyper-parameters $\alpha$ (in Eq. \ref{tripletloss}) is empirically set to 0.20. We set the hyper-parameters $\beta$ (in Eq. \ref{supervision2}) and $\gamma$ (in Eq. \ref{part_kl_label}) to 0.25 and 0.20, respectively. The number of body parts, i.e. $K$, is set to 6, according to the evaluation results presented in Fig. \ref{K}.

BCD-Net is implemented using the PyTorch framework. The standard stochastic gradient descent (SGD) optimizer with a weight decay of $5 \times 10^{-4}$ and a momentum \cite{sutskever2013importance} value of 0.9 is utilized for model optimization. Fine-tuned from the IDE model \cite{zheng2017person}, the BCD-Net is trained in an end-to-end fashion for $70$ epochs. The learning rate is initially set to 0.01 and then multiplied by 0.1 every 20 epochs.

\subsection{Ablation Study}
\newcommand{\tabincell}[2]{\begin{tabular}{@{}#1@{}}#2\end{tabular}}
\newcommand{\thickhline}{
    \noalign {\ifnum 0=`}\fi \hrule height 1pt
    \futurelet \reserved@a \@xhline}

\begin{table*}[t]
\caption{Ablation Study on Each Key Component of BCD-Net}
\centering
\begin{center}
\begin{tabular}{p{1.8cm}<{\centering}|p{1.25cm}<{\centering} p{0.4cm}<{\centering} p{0.4cm}<{\centering} p{0.4cm}<{\centering} p{0.40cm}<{\centering}|cc|cc|cc|cc}
  \hline
  Dataset & \multicolumn{5}{c|}{Components} & \multicolumn{2}{c|}{Market-1501} & \multicolumn{2}{c|}{MSMT17} &\multicolumn{2}{c|}{CUHK03-Detected} &\multicolumn{2}{c}{CUHK03-Labeled} \\

  \hline
  Metric &Extra Nets &CA &$\mathcal{L}_{C}^k$ &$\mathcal{L}_{p}^k$ &$\mathcal{L}_{h}$ & Rank-1  & mAP  & Rank-1 & mAP & Rank-1 & mAP & Rank-1 & mAP \\
  \hline
  \hline
  Baseline                       &-           &-          &-          &-          &-          &94.4&85.6      &77.3&53.5   &74.1&69.3   &78.7&73.9\\
  Baseline+                      &\checkmark  &-          &-          &-          &-          &94.7&87.1      &80.9&58.3   &76.8&72.0   &80.4&76.5\\
  \hline
  \hline
  \multirow{2}*{\tabincell{c}{BCCA}}
                                 &\checkmark  &\checkmark &-          &-          &-          &94.9&87.9      &81.6&59.2   &78.8&75.1   &81.2&77.8\\
                                 &\checkmark  &\checkmark &\checkmark &-          &-          &95.7&88.5      &83.5&61.9   &83.1&77.9   &85.1&80.6\\
  \hline
  \hline
  \multirow{3}*{\tabincell{c}{Spatial \\ Regularizations}}
                                 &\checkmark  &-          &-          &\checkmark &-          &95.5&88.5      &82.6&60.9   &81.3&76.6   &83.1&79.4\\
                                 &\checkmark  &-          &-          &-          &\checkmark &95.2&87.9      &81.9&60.2   &80.6&76.1   &82.1&78.3\\
                                 &\checkmark  &-          &-          &\checkmark &\checkmark &95.7&88.7      &83.1&61.9   &82.0&76.9   &84.3&79.7\\
  \hline
  \hline
  BCD-Net                        &\checkmark  &\checkmark &\checkmark &\checkmark &\checkmark &96.2&89.5      &84.1&63.7   &84.2&78.7   &86.2&81.6\\
  \hline
\end{tabular}
\end{center}
\label{Ablation1}
\end{table*}

We first validate the effectiveness of each newly introduced component in BCD-Net, i.e., BCCA and the pair of spatial regularization terms. Experiments are conducted on the Market-1501, MSMT17, and CUHK03 datasets. The results are tabulated in Table \ref{Ablation1}.

In Table \ref{Ablation1}, ``Baseline" denotes our baseline model, which is constructed by removing both BCCA and the spatial regularization terms from BCD-Net illustrated in Fig. \ref{framework}. Moreover, ``Baseline+" builds $K$ extra sub-networks on the backbone model compared to ``Baseline" in the training stage. These extra sub-networks are removed during testing. They have the same structure as the sub-network illustrated in Fig. \ref{framework}. The way in which input feature maps are obtained for the $K$ extra sub-networks is the same as that used in the Part-based Convolutional Baseline model \cite{ECCVsun2018beyond}. In brief, we uniformly slice $\mathbf{T}_{i}$ in the height dimension into $K$ smaller feature maps, which form the input of these extra sub-networks. As shown in Table \ref{Ablation1}, ``Baseline+" outperforms the baseline model; this is because the extra sub-networks play a regularization role, which encourages the backbone model to extract more diverse and fine-grained features. To improve the performance of BCD-Net, we include the $K$ extra sub-networks for all models except ``Baseline" in Table \ref{Ablation1}.

\subsubsection{Effectiveness of BCCA}
In this experiment, we demonstrate the effectiveness of BCCA. In Table \ref{Ablation1}, we equip the ``Baseline+" model with plain CA modules (without supervision) and the supervision $\mathcal{L}_{C}^k$ in Eq. \ref{bcca}, successively. The experimental results presented in Table \ref{Ablation1} show that plain CA modules slightly promote the performance of the ``Baseline+" model. For example, performance improvements of 0.2\% and 0.8\% can be observed on Market-1501 in terms of Rank-1 accuracy and mAP, respectively. Moreover, further applying the supervision $\mathcal{L}_{C}^k$ results in the entire BCCA module obtaining a considerable performance promotion on all three benchmarks compared with ``Baseline+": in brief, Rank-1 accuracy is improved by 1.0\%, 2.6\%, 6.3\%, and 4.7\% and mAP by 1.4\%, 3.6\%, 5.9\%, and 4.1\% on each dataset, respectively. These experimental results demonstrate the effectiveness of BCCA.

\begin{figure}[t]
    \centerline{\includegraphics[width=0.495\textwidth]{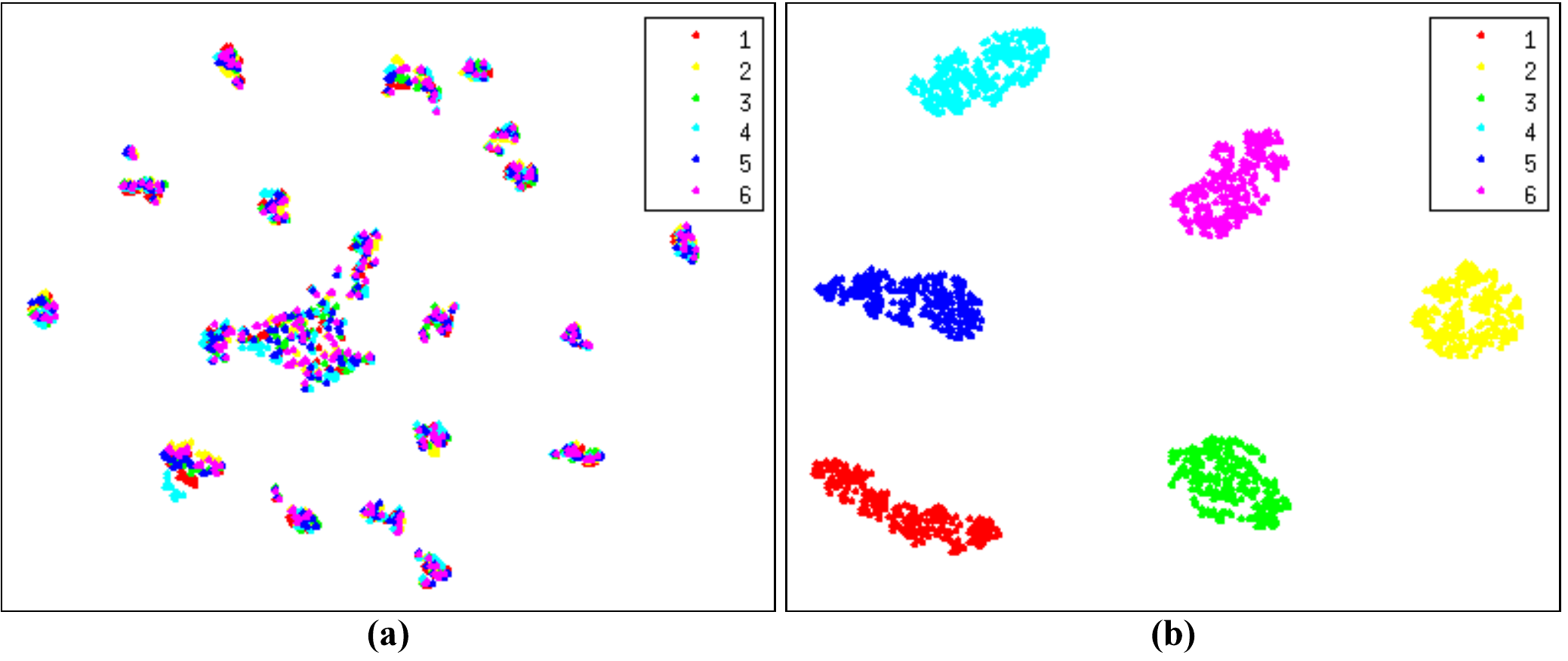}}
     \caption{Visualization using t-SNE \cite{maaten2008visualizing} for the channel weights produced by $K$ plain CA modules and $K$ BCCAs, respectively. (a) visualization for plain CA modules; (b) visualization for BCCAs. We sample 10 images for each of 25 random identities. $K$ body parts are represented using different colors. In this figure, $K$ is equal to 6.}
    \label{visualization_distribution}
\end{figure}

Moreover, we adopt t-SNE \cite{maaten2008visualizing} to visualize the channel weights produced by plain CA modules and BCCAs in Fig. \ref{visualization_distribution}, respectively. As shown in Fig. \ref{visualization_distribution}(a), the channel weights learned by the $K$ plain CA modules tend to be similar for each image, meaning that the plain CA modules cannot help BCD-Net to learn diverse part-level features. By contrast, the output of $K$ BCCAs for each image are diverse, as illustrated in Fig. \ref{visualization_distribution}(b). In addition, the outputs of the same BCCA module tend to be similar for different images. This indicates that BCCAs capture the general characteristic of each body part across different identities. The above analysis reveals the effectiveness of the proposed supervision signal that drives BCD-Net to learn diverse part-aware features.

\subsubsection{Effectiveness of Spatial Regularizations}
In this experiment, we demonstrate the effectiveness of the pair of spatial regularization terms. Table \ref{Ablation1} presents the results of equipping the ``Baseline+" model with the part-level regularization term ($\mathcal{L}_{p}^k$ in Eq. \ref{part_spatial}), the holistic-level regularization term ($\mathcal{L}_{h}$ in Eq. \ref{holistic_spatial}), and both terms, respectively. After assessing these results, we can make the following observations. Firstly, equipping either the part- or holistic-level regularization consistently brings about performance gains. In particular, the part-level regularization improves Rank-1 accuracy by 0.8\% and 1.7\% on Market-1501 and MSMT17, respectively. Secondly, the combination of part- and holistic-level regularizations provides a further accuracy boost, suggesting that complementarity exists between the two regularizations. Finally, the combination of the two kinds of spatial regularizations improve the performance of ``Baseline+" by 1.0\%, 2.2\%, 5.2\%, and 3.9\% in terms of Rank-1 accuracy, as well as by 1.6\%, 3.6\%, 4.9\%, and 3.2\% in terms of mAP on each dataset, respectively. These results demonstrate the effectiveness of employing the pair of spatial regularizations.

\subsubsection{Combination of BCCA and Spatial Regularizations}
In this experiment, we equip ``Baseline+" with both BCCA and the pair of spatial regularization terms; this model is referred to as BCD-Net in Table \ref{Ablation1}. We can observe that BCD-Net consistently outperforms all other comparison models in Table \ref{Ablation1}. This result indicates that BCCA and the pair of spatial regularization terms are complementary in promoting the learning of semantically aligned part-level features. Finally, BCD-Net outperforms ``Baseline+" by 1.5\%, 3.2\%, 7.4\%, and 5.8\% in terms of Rank-1 accuracy, and by 2.4\%, 5.4\%, 6.7\%, and 5.1\% in terms of mAP on each dataset, respectively. The above experimental results demonstrate the effectiveness of BCD-Net.

\subsubsection{Visualization of Attention Maps for BCD-Net}
\begin{figure}[t]
    \centerline{\includegraphics[width=0.50\textwidth]{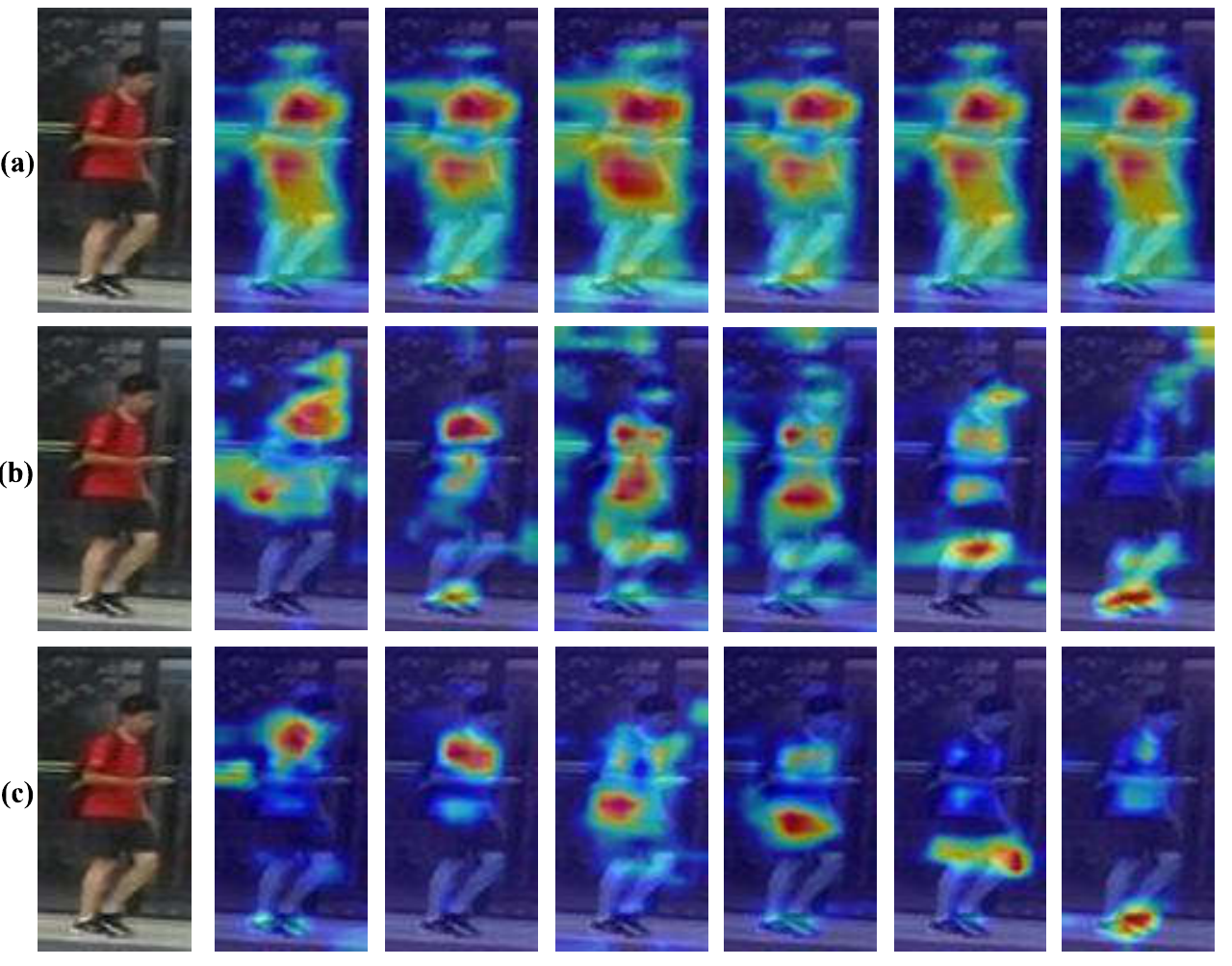}}
     \caption{Visualization of attention maps for each of the $K$ sub-network's classifier using Grad-CAM \cite{selvaraju2017grad}. Three representative models in Table \ref{Ablation1} are compared: (a) ``Baseline+"; (b) ``Baseline+" with BCCA; (c) BCD-Net. (Best viewed in color.)}
    \label{visualization1}
\end{figure}

We further support the above experimental results by visualizing the attention maps for the predictions of each of the $K$ classifiers using Grad-CAM \cite{selvaraju2017grad}. Three representative models are compared in Table \ref{Ablation1}: ``Baseline+", ``Baseline+" with BCCA, and BCD-Net. The attention maps of one pedestrian image with the misalignment problem are illustrated in Fig. \ref{visualization1}. As can be seen in Fig. \ref{visualization1}(a), the $K$ attention maps of ``Baseline+" are similar and consistently highlight the most discriminative regions of the pedestrian. This is because all sub-networks in ``Baseline+" share exactly the same input feature maps and are optimized without any part-specific supervision signals; therefore, their output lacks diversity. By contrast, BCCA encourages these sub-networks to learn diverse part-aware features, as shown in Fig. \ref{visualization1}(b). Moreover, the inclusion of the pair of spatial regularization terms improves the quality of the part features by encouraging them to be more spatially attentive for each respective part, as illustrated in Fig. \ref{visualization1}(c).

\begin{figure}[t]
    \centerline{\includegraphics[width=0.5\textwidth]{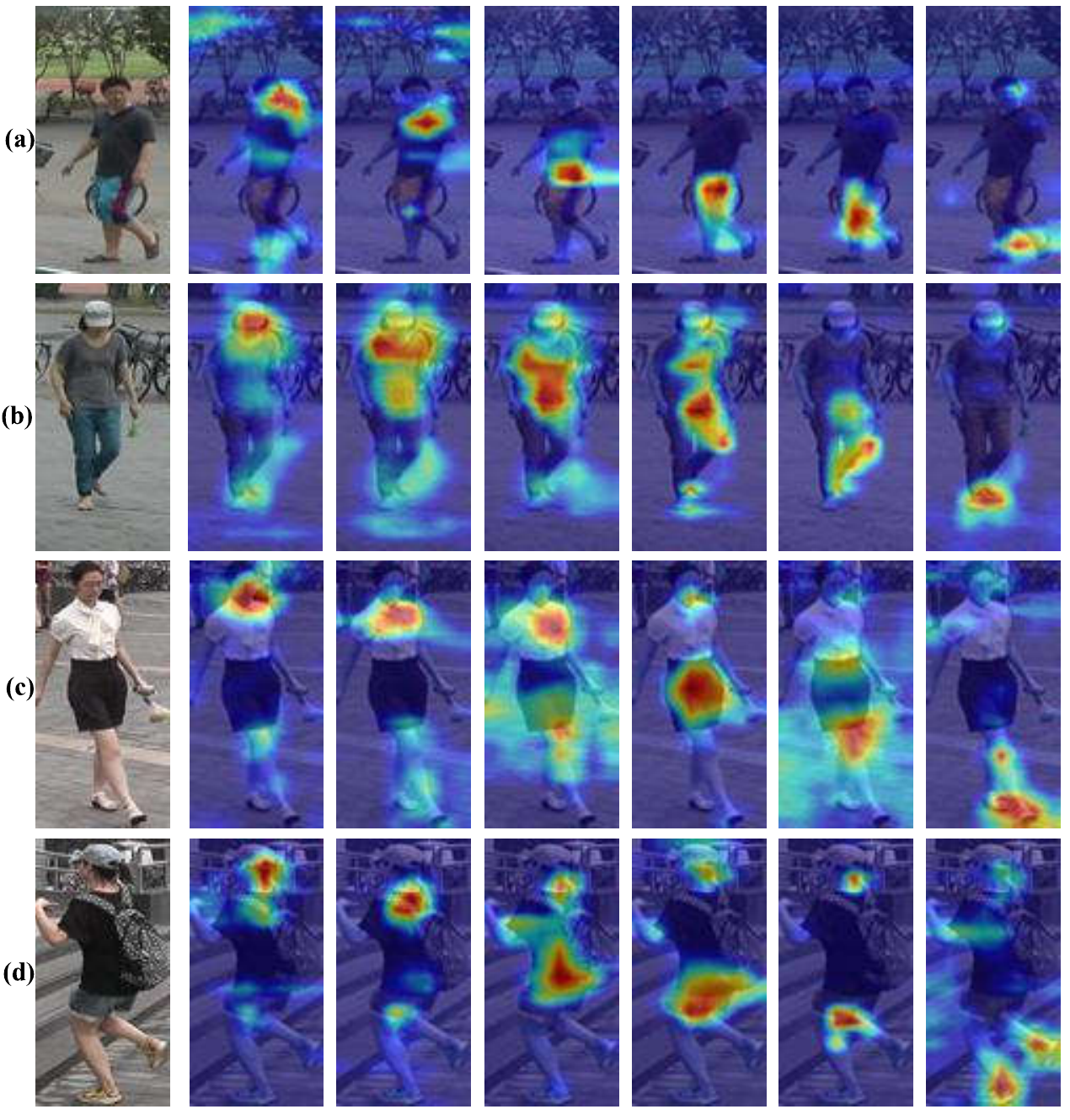}}
     \caption{Visualization of attention maps for each of the $K$ sub-network's classifier using Grad-CAM \cite{selvaraju2017grad}. The attention maps of the same sub-network are semantically consistent across images, even in the face of errors in pedestrian detection (a, b) and pose variations (c, d). (Best viewed in color.)}
    \label{visualization2}
\end{figure}

In addition, we further visualize the attention maps generated by BCD-Net on more pedestrian images in Fig. \ref{visualization2}. It can be seen from these visualizations that BCD-Net attends adaptively to the body parts, even when alignment errors in pedestrian detection (Fig. \ref{visualization2}(a, b)) or changes of pose (Fig. \ref{visualization2}(c, d)) are present. The above visualization results demonstrate the effectiveness of BCD-Net.

\subsubsection{Evaluation on the Value of Hyper-parameters $\beta$ and $\gamma$}
In this experiment, we evaluate the value of two important hyper-parameters, i.e., $\beta$ (in Eq. \ref{supervision2}) and $\gamma$ (in Eq. \ref{part_kl_label}). To facilitate clean comparison, the ``Baseline+" is only equipped with the proposed BCCA and the pair of spatial regularizations for two hyper-parameters, respectively. The other experimental settings remain unchanged.

\begin{figure}[t]
    \centerline{\includegraphics[width=0.495\textwidth]{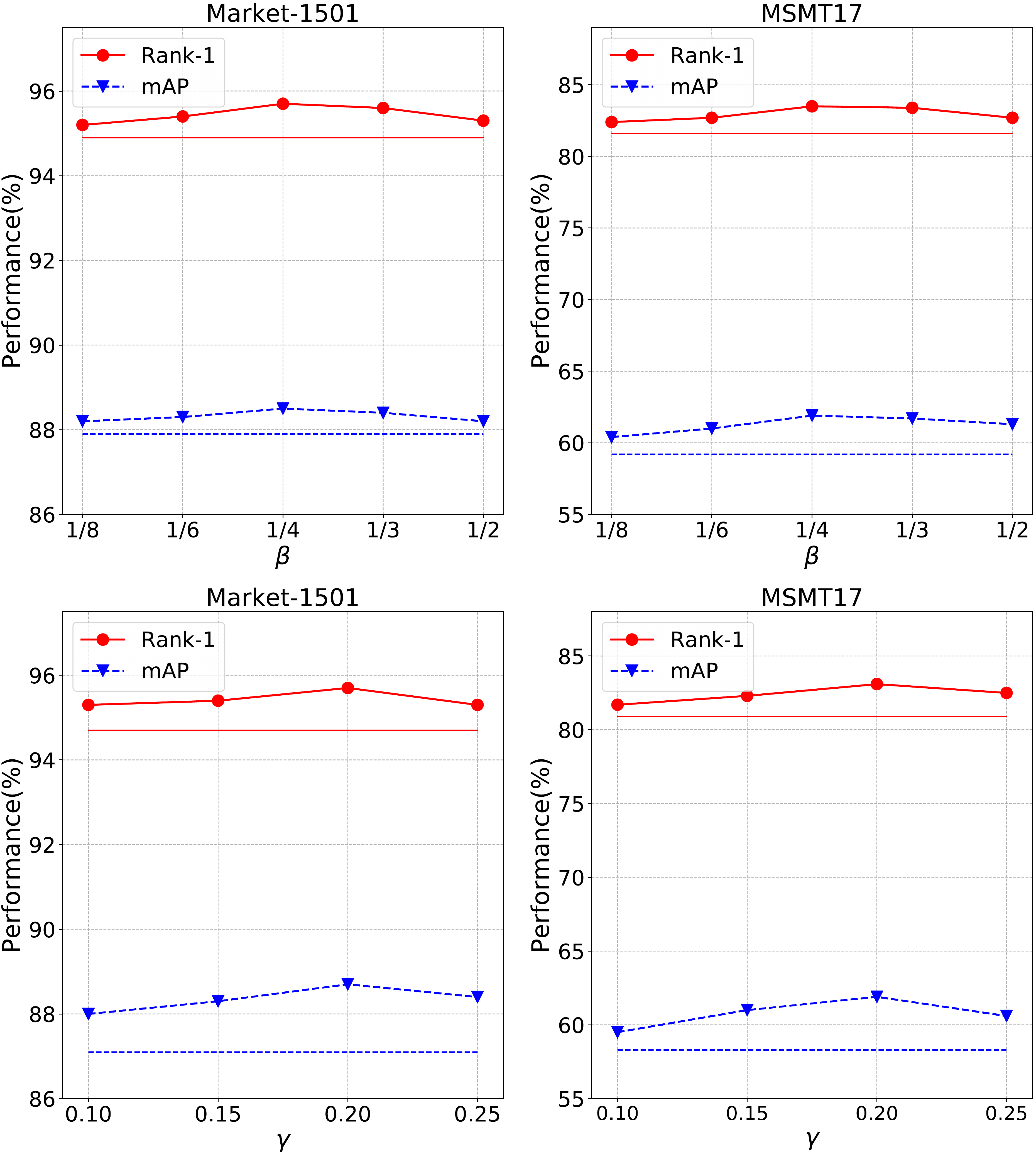}}
     \caption{Evaluation on the value of hyper-parameters $\beta$ and $\gamma$. The straight lines in the two figures for $\beta$ represent the performance of ``Baseline+ with plain CAs". While the straight lines in the two figures for $\gamma$ represent the performance of ``Baseline+".}
    \label{betagamma}
\end{figure}

Experimental results are illustrated in Fig. \ref{betagamma}. We can make the following observations. First, the performance becomes better when the value of $\beta$ or $\gamma$ increases to a certain extent. Second, the performance drops when the value of $\beta$ or $\gamma$ further increases. For example, the performance becomes better when the value of $\beta$ increases from $\frac{1}{8}$ to $\frac{1}{4}$. This is because a moderately large $\beta$ is helpful to robustly select part-relevant channels for each body part. Besides, the performance drops when the value of $\beta$ further increases from $\frac{1}{4}$ to $\frac{1}{2}$. This may be because when the value of $\beta$ is too large, many weakly relevant channels to each part are ignored, which lowers the quality of part features. 

\subsection{Comparisons with Variants of BCD-Net}
\subsubsection{Comparisons with Variants for the Supervision Signal $\hat{\mathbf{C}}^{k}$}
We compare the performance of the proposed supervision signal for BCCA (i.e., $\hat{\mathbf{C}}^{k}$) with two possible variants, with results presented in Table \ref{Ablation2}. The first variant, denoted as ``One-hot Vector", formulates $\mathbf{\widetilde{v}}_{c}$ as a one-hot label vector. In brief, we set the largest element in $\mathbf{\widetilde{v}}_{c}$ to 1 and the others to 0; this supervision signal encourages each channel to describe one exclusive body part. The second variant, denoted as ``No Filtration", removes the channel filtration step in Eq. \ref{supervision2}. This supervision allows all channels to be used for part-level feature extraction. To facilitate clean comparison, we remove the pair of spatial regularization terms in BCD-Net and evaluate the BCCA module with each of the three supervision signals.

An examination of Table \ref{Ablation2} allows us to make the following observations. First, all three types of supervision signals boost the ReID performance; this further proves that the batch-coherence based supervision signals are effective. Second, both variants achieve inferior performance relative to our proposed approach. For example, our proposed method outperforms the two variants by 1.1\% and 0.7\% respectively in terms of Rank-1 accuracy on the MSMT17 database. The above comparisons demonstrate the superiority of the proposed supervision signal in BCCA.

\begin{table}
\caption{Performance Comparison with Variants of the Supervision Signal for BCCA}
\centering
\begin{tabular}{c|cc|cc}
\hline
  Dataset & \multicolumn{2}{c|}{Market-1501} & \multicolumn{2}{c}{MSMT17} \\
  \hline
  Metric  & Rank-1  & mAP  & Rank-1 & mAP \\
  \hline
  \hline
  Baseline+         &94.7&87.1 &80.9&58.3 \\
  \hline

  One-hot Vector    &95.4&88.2&82.4&61.0 \\
  No Filtration     &95.5&88.4&82.8&61.2 \\
  Ours              &95.7&88.5&83.5&61.9 \\
  \hline
\end{tabular}
\label{Ablation2}
\end{table}

\subsubsection{Comparisons with Variants for BCCA}
Next, we compare BCCA with two possible variants in Table \ref{Ablation3}. The first of these variants penalizes the cosine distance between $\hat{\mathbf{C}}^{k}$ and the output of CA modules for each respective image. The second variant applies the supervision $\hat{\mathbf{C}}^{k}$ in Eq. \ref{bcca} during training in the same way as BCCA. However, it utilizes $\bar{\mathbf{C}}^{k}$ rather than $\mathbf{C}^{k}_{i}$ to perform channel-wise multiplication with $\mathbf{T}_i$ in Eq. \ref{multiplication}. Therefore, this method adopts the same channel attention for a batch of images. The two variants are denoted as ``Variant 1" and ``Variant 2" respectively in Table \ref{Ablation3}.

As can be seen from Table \ref{Ablation3}, the performance of both variants is inferior to that of our proposed BCCA. For example, BCCA outperforms the two variants by 0.9\% and 1.5\% respectively in terms of mAP on the MSMT17 dataset. These results indicate that the output of BCCA should be adjustable for each image, although the images follow similar part-channel correspondence. The above comparisons demonstrate the superiority of the proposed BCCA module.

\begin{table}
\caption{Performance Comparison with Variants for BCCA}
\centering
\begin{tabular}{c|cc|cc}
\hline
  Dataset & \multicolumn{2}{c|}{Market-1501} & \multicolumn{2}{c}{MSMT17} \\
  \hline
  Metric  & Rank-1  & mAP  & Rank-1 & mAP \\
  \hline
  \hline
  Baseline+             &94.7 &87.1 &80.9 &58.3 \\
  \hline
  Variant1              &95.4 &88.4 &82.5 &61.0 \\
  Variant2              &95.3 &88.1 &82.2 &60.4 \\
  Ours                  &95.7 &88.5 &83.5 &61.9 \\
  \hline
\end{tabular}
\label{Ablation3}
\end{table}

\subsubsection{Comparisons with Variant for the Supervision Signal $\bold{\hat{z}}^{k}(l)$}
In this experiment, we compare the performance of the supervision signal $\bold{\hat{z}}^{k}(l)$ in Eq. \ref{part_kl_label} for part-level spatial regularization with one possible variant. In Eq. \ref{part_kl_label}, we assign a small value to those elements in $\bold{\hat{z}}^{k}(l)$ that do not correspond to the $k$-th part. This setting is designed to account for the variation of body part locations. One natural alternative would be to impose a more rigid constraint: in brief, this involves setting the value of the $\frac{H}{K}$ elements that correspond to the $k$-th part in $\bold{\hat{z}}^{k}(l)$ to $\frac{K}{H}$, while the value of all other elements is set to 0. We refer to this variant as ``Hard Label" in Table \ref{Ablation4}. To facilitate clean comparison, we equip ``Baseline+" only with the pair of spatial regularizations, but impose different supervision signals for the part-level regularization term.

Experimental results are tabulated in Table \ref{Ablation4}. These results indicate that the proposed supervision signal $\bold{\hat{z}}^{k}(l)$ outperforms the ``Hard Label" variant by a noticeable margin. For example, $\bold{\hat{z}}^{k}(l)$ outperforms the variant by 0.3\% and 0.7\% in terms of Rank-1 accuracy on Market-1501 and MSMT17, respectively. The above comparisons demonstrate that it is necessary to account for the change of part locations in $\bold{\hat{z}}^{k}(l)$ due to the errors in pedestrian detection and pose variations.

\begin{table}
\caption{Performance Comparison with One Variant for Part-level Spatial Regularization}
\centering
\begin{tabular}{c|cc|cc}
\hline
  Dataset & \multicolumn{2}{c|}{Market-1501} & \multicolumn{2}{c}{MSMT17} \\
  \hline
  Metric  & Rank-1  & mAP  & Rank-1 & mAP \\
  \hline
  \hline
  Baseline+   &94.7&87.1 &80.9&58.3 \\
  \hline
  Hard Label  &95.4&88.4&82.4 &60.6 \\
  Ours        &95.7&88.7&83.1 &61.9 \\
  \hline
\end{tabular}
\label{Ablation4}
\end{table}

\subsection{Comparisons with State-of-the-Art Methods}
We next compare the performance of BCD-Net with that of state-of-the-art methods on four large-scale ReID benchmarks: namely, Market-1501 \cite{zheng2015scalable}, DukeMTMC-reID \cite{zheng2017unlabeled}, CUHK03 \cite{li2014deepreid}, and MSMT17 \cite{wei2018person}. Moreover, to facilitate fair comparison, the existing approaches are sorted into two categories: holistic feature-based (HF) methods and part feature-based (PF) methods. The PF methods can be further divided into two sub-categories: methods that extract part-level features from fixed locations (PF-fixed) and methods that learn part-aware features (PF-aware).

\subsubsection{Performance Comparisons on Market-1501}
\begin{table}[t]
\caption{Performance Comparisons on the Market-1501 Dataset}
\centering
\begin{tabular}{c|c|cc|cc}
\hline
\multicolumn{2}{c|}{\multirow{2}{*}{Methods}} & \multicolumn{2}{c|}{Single Query} & \multicolumn{2}{c}{Multiple Query} \\ \cline{3-6}
  \multicolumn{2}{c|}{} &Rank-1  & mAP  & Rank-1 & mAP \\
\hline
\hline
 \multirow{11}*{\rotatebox{90}{HF}}
  &PSE \cite{CVPRsaquib2018pose} &87.7 &69.0 &- &- \\
  &MLFN \cite{CVPRchang2018multi} &90.0 &73.4 &92.3 &82.4 \\
  &DuATM \cite{CVPRsi2018dual} &91.4 &76.6 &- &- \\
  &SFT \cite{ICCVLuo2019Spectral} &93.4 &82.7 & - & - \\
  &IANet \cite{CVPRhou2019interaction} &94.4 &83.1 & - & - \\
  &OSNet \cite{ICCVZhou2019Omni} &94.8 &84.9 &- & - \\
  &BDB-Cut \cite{ICCVDai2019Batch} &95.3 &86.7 &- & - \\
  &ABDNet \cite{ICCVChen2019ABD-Net} &95.6 &88.3 &- & - \\
  &SCSN(3 stages) \cite{chen2020salience} &95.7 &88.5 &- & - \\
  &SCAL(channel) \cite{ICCVChen2019Self} &95.8 &89.3 &- &- \\
  &SAN \cite{AAAIxin2020semantics} &96.1 &88.0 &- &- \\
\hline
\hline
 \multirow{5}*{\rotatebox{90}{PF-fixed}}
  &PCB \cite{ECCVsun2018beyond} & 92.3 &77.4 & - & - \\
  &HPM \cite{fu2019horizontal} &94.2 &82.7 &- &- \\
  &MHN-6 (PCB)\cite{ICCVChen2019Mixed} &95.1 &85.0 & - & -\\
  &MGN \cite{wang2018learning} &95.7 &86.9 &96.9 &90.7 \\
  &Pyramid \cite{CVPRzheng2019pyramidal} &95.7 &88.2 & - & - \\
\hline
\hline
 \multirow{14}*{\rotatebox{90}{PF-aware}}
  &MSCAN \cite{CVPRli2017learning} &80.3 &57.5 & 86.8 &66.7 \\
  &PAR \cite{ICCVzhao2017deeply} &81.0 &63.4 & - & - \\
  &PDC \cite{ICCVsu2017pose} &84.1 &63.4 & - & - \\
  &AACN \cite{CVPRxu2018attention} & 85.9 &66.9 & 89.8 &75.1 \\
  &PL-NET \cite{TIPyao2019deep} &88.2 &69.3 & & \\
  &HA-CNN \cite{CVPRli2018harmonious} &91.2 &75.7 &93.8 &82.8 \\
  &Part-Aligned \cite{ECCVsuh2018part} & 91.7 & 79.6 & 94.0 & 85.2 \\
  &PCB+RPP \cite{ECCVsun2018beyond} & 93.8 &81.6 & - & - \\
  &BAT-net \cite{ICCVFang2019Bilinear} &95.1 &81.4 &- &- \\
  &ISP \cite{zhu2020identity} &95.3 & 88.6 & - & - \\
  &FPR \cite{ICCVHe2019Foreground} &95.4 &86.6 & - & - \\
  &DSA-reID \cite{CVPRzhang2019densely} & 95.7 &87.6 & - & - \\
  &CDPM \cite{wang2019cdpm} &95.9 &87.2 &97.0 &91.1 \\
  &\bfseries BCD-Net &{\bfseries 96.2} &{\bfseries 89.5} &{\bfseries 97.0} &{\bfseries 92.7} \\
\hline
\end{tabular}
\label{1501}
\end{table}
The performance of BCD-Net and the state-of-the-art methods on Market-1501 are tabulated in Table \ref{1501}. After examining the results, we can make the following three observations.

First, BCD-Net outperforms all HF methods. When compared with one of the most recent HF methods, i.e. SAN \cite{AAAIxin2020semantics}, BCD-Net still exhibits a noticeable performance gain of 1.5\% in terms of mAP under the single-query mode. Moreover, BCD-Net is easier to use than SAN; this is because SAN relies on a 3D model to normalize each training image. Besides, SAN adopts a stage-wise strategy during training. By contrast, BCD-Net can be trained using the standard end-to-end strategy in a single stage.

Second, BCD-Net outperforms all PF-fixed methods. For example, BCD-Net outperforms MGN \cite{wang2018learning} by 0.5\% and 2.6\% in terms of the Rank-1 accuracy and mAP respectively under the single-query mode. Another advantage of BCD-Net is that it extracts single-scale features, while MGN extracts multi-scale part-level features; therefore, BCD-Net is more efficient.

Third, BCD-Net surpasses all PF-aware methods. For example, BCD-Net beats DSA-reID \cite{CVPRzhang2019densely} by 0.5\% in terms of Rank-1 accuracy and 1.9\% in terms of mAP under the single-query mode. In addition, BCD-Net does not require part detection during either training or testing, while DSA-reID depends on precise part detection during training; as a result, BCD-Net is easier to use in practice. These findings indicate that BCD-Net is effective in extracting high-quality part-aware representations.

\subsubsection{Performance Comparisons on DukeMTMC-ReID}
Comparisons on the DukeMTMC-ReID dataset are summarized in Table \ref{duke}. These results show that BCD-Net outperforms all other approaches by a significant margin in terms of both Rank-1 accuracy and mAP. In particular, BCD-Net beats one of the most recent part-based methods, i.e., MHN-6(PCB) \cite{ICCVChen2019Mixed}, by 2.0\% and 4.4\% in terms of Rank-1 accuracy and mAP respectively. These comparisons further demonstrate the effectiveness of BCD-Net.

\begin{table}[h]
\centering\caption{Performance Comparisons on The DukeMTMC-ReID Dataset}
\centering
\begin{tabular}{c|c|cc}
\hline
  \multicolumn{2}{c|}{Methods} & Rank-1 & mAP \\
\hline
\hline
 \multirow{9}*{\rotatebox{90}{HF}}
  & DaRe \cite{CVPRwang2018resource} &75.2 & 57.4\\
  & SVDNet \cite{ICCVsun2017svdnet} &76.7 & 56.8\\
  & PSE \cite{CVPRsaquib2018pose} &79.8 &62.0  \\
  & DuATM \cite{CVPRsi2018dual} &81.8 &64.6 \\
  & Mancus \cite{ECCVwang2018mancs} & 84.9 &71.8 \\
  & SFT \cite{ICCVLuo2019Spectral} &86.9 &73.2 \\
  & IANet \cite{CVPRhou2019interaction} &87.1 &73.4 \\
  & SAN \cite{AAAIxin2020semantics} &87.9 &75.5 \\
  & OSNet \cite{ICCVZhou2019Omni} &88.6 &73.5 \\
\hline
\hline
  \multirow{5}*{\rotatebox{90}{PF-fixed}}
  & PCB \cite{ECCVsun2018beyond} & 83.3 & 69.2 \\
  & HPM \cite{fu2019horizontal} & 86.6 & 74.3 \\
  & MGN \cite{wang2018learning} & 88.7 &78.4 \\
  & Pyramid \cite{CVPRzheng2019pyramidal} &89.0 &79.0 \\
  & MHN-6 (PCB)\cite{ICCVChen2019Mixed} &89.1 &77.2 \\
\hline
\hline
  \multirow{9}*{\rotatebox{90}{PF-aware}}
  & AACN \cite{CVPRxu2018attention} & 76.8 & 59.3\\
  & HA-CNN \cite{CVPRli2018harmonious} &80.5 & 63.8\\
  & Part-aligned \cite{ECCVsuh2018part} &84.4 & 69.3\\
  & DSA-reID \cite{CVPRzhang2019densely} & 86.2 &74.3 \\
  & DG-Net \cite{CVPRzheng2019joint} &86.6 &74.8 \\
  & BAT-net \cite{ICCVFang2019Bilinear} &87.7 &77.3 \\
  & MuDeep \cite{qian2019leader} &88.2 &75.6 \\
  & CDPM \cite{wang2019cdpm} &90.1 &80.2 \\
  &\bfseries BCD-Net &{\bfseries 91.1} &{\bfseries 81.6} \\
\hline
\end{tabular}
\label{duke}
\end{table}

\subsubsection{Performance Comparisons on CUHK03}
We next compare the performance of BCD-Net with state-of-the-art methods on the CUHK03 dataset in Table \ref{cuhk03}. Both the manually labelled and auto-detected bounding boxes of CUHK03 are utilized to evaluate the effectiveness of BCD-Net.

The comparisons in Table \ref{cuhk03} reveal that BCD-Net again achieves the best results, outperforming state-of-the-art methods with clear margins. For example, BCD-Net scores 86.2\% and 81.6\% on Rank-1 accuracy and mAP respectively on CUHK03-Labeled; this represents an improvement over Pyramid \cite{CVPRzheng2019pyramidal} of 7.3\% in terms of Rank-1 accuracy and 4.7\% in terms of mAP. Moreover, it is worth noting that Pyramid extracts multi-scale part-level features, while BCD-Net only extracts single-scale features. In addition, when compared with one recent PF-aware method, i.e., DSA-reID \cite{CVPRzhang2019densely}, BCD-Net also exhibits obvious advantages. For example, BCD-Net outperforms DSA-reID by 6.0\% in terms of Rank-1 accuracy and 5.6\% in terms of mAP on CUHK03-Detected. The above comparisons demonstrate the effectiveness of BCD-Net.

\begin{table}[t]
\centering
\caption{Performance Comparisons on The CUHK03 Dataset}
\begin{tabular}{c|c|cc|cc}
\hline
  \multicolumn{2}{c|}{\multirow{2}{*}{Methods}} & \multicolumn{2}{c|}{Detected} & \multicolumn{2}{c}{Labeled} \\ \cline{3-6}
  \multicolumn{2}{c|}{} &Rank-1  & mAP  & Rank-1 & mAP \\
\hline
\hline
  \multirow{5}*{\rotatebox{90}{HF}}
  & SVDNet \cite{ICCVsun2017svdnet}  &41.5 &37.3  &40.9 &37.8 \\
  & Mancus \cite{ECCVwang2018mancs} & 65.5 &60.5 &69.0 &63.9 \\
  & OSNet \cite{ICCVZhou2019Omni} &72.3 &67.8 & - & - \\
  & BDB+Cut \cite{ICCVDai2019Batch} &76.4 &73.5 &- & - \\
  & SAN \cite{AAAIxin2020semantics} &79.4 &74.6 &80.1 &76.4 \\
\hline
\hline
  \multirow{5}*{\rotatebox{90}{PF-fixed}}
  & PCB \cite{ECCVsun2018beyond} &61.3 &54.2 & - & - \\
  & HPM \cite{fu2019horizontal}  &63.9 &57.5 & - & - \\
  & MGN \cite{wang2018learning}  &66.8 &66.0 &68.0 &67.4 \\
  & MHN-6 (PCB)\cite{ICCVChen2019Mixed} &71.7 &65.4 &77.2 &72.4\\
  & Pyramid \cite{CVPRzheng2019pyramidal} &78.9 &74.8 &78.9 &76.9 \\
\hline
\hline
\multirow{10}*{\rotatebox{90}{PF-aware}}
  & HA-CNN \cite{CVPRli2018harmonious} &41.7 &38.6 &44.4 &41.0 \\
  & AACN \cite{CVPRxu2018attention}  &46.7 &46.9  &50.1 &50.2 \\
  & MLFN \cite{CVPRchang2018multi} &52.8 &47.8 &54.7 &49.2 \\
  & PCB+RPP \cite{ECCVsun2018beyond} &63.7 &57.5 & - & - \\
  &ISP \cite{zhu2020identity} &75.2 &71.4 &76.5 &74.1 \\
  & MuDeep \cite{qian2019leader} &71.9 &67.2 &75.6 &70.5\\
  & BAT-net \cite{ICCVFang2019Bilinear} &76.2 &73.2  &78.6 &76.1 \\
  & DSA-reID \cite{CVPRzhang2019densely} &78.2 &73.1 &78.9 &75.2  \\
  & CDPM \cite{wang2019cdpm} &78.8 &73.3 &81.4 &77.5
  \\
  &\bfseries BCD-Net &{\bfseries 84.2} &{\bfseries 78.7} &{\bfseries 86.2} &{\bfseries 81.6}\\
\hline
\end{tabular}
\label{cuhk03}
\end{table}

\subsubsection{Performance Comparisons on MSMT17}
\begin{table}[t]
\centering\caption{Performance Comparisons on The MSMT17 Dataset}
\centering
\begin{tabular}{c|c|cc}
\hline
  \multicolumn{2}{c|}{Methods} & Rank-1 & mAP \\
\hline
\hline
  \multirow{7}*{\rotatebox{90}{HF}}
  & PGR \cite{li2019pose} &66.0 &37.9 \\
  & SFT \cite{ICCVLuo2019Spectral} &73.6 &47.6 \\
  & IANet \cite{CVPRhou2019interaction} &75.5 &46.8 \\
  & DG-Net \cite{CVPRzheng2019joint} &77.2 &52.3 \\
  & OSNet \cite{ICCVZhou2019Omni} & 78.7 & 52.9 \\
  & RGA-SC \cite{RAGA} &81.3 &56.3 \\
  & ABDNet \cite{ICCVChen2019ABD-Net} &82.3 &60.8 \\
\hline
\hline
  \multirow{5}*{\rotatebox{90}{PF}}
  & PDC \cite{ICCVsu2017pose, PTGAN} & 58.0 &29.7  \\
  & GLAD \cite{glad, PTGAN} &61.4 &34.0 \\
  & PCB+RPP \cite{ECCVsun2018beyond, CVPRzheng2019joint} & 68.2 &40.4  \\
  & BAT-net \cite{ICCVFang2019Bilinear} &79.5 &56.8 \\
  & \bfseries BCD-Net &{\bfseries 84.1} &{\bfseries 63.7} \\
\hline
\end{tabular}
\label{msmt17}
\end{table}
Finally, we compare the performance of BCD-Net with state-of-the-art approaches on MSMT17 in Table \ref{msmt17}. As this dataset was only recently released, only a limited number of works have reported results on this dataset. We categorize existing these approaches into two groups, i.e. methods based on holistic- and part-level features.

The results of the comparisons in Table \ref{msmt17} reveal that BCD-Net achieves both the best Rank-1 accuracy and the best mAP. For example, BCD-Net outperforms one of the most recent HF methods, ABDNet \cite{ICCVChen2019ABD-Net}, by 1.8\% and 2.9\% in terms of Rank-1 accuracy and mAP respectively. Moreover, BCD-Net also significantly outperforms existing part-based approaches. Finally, BCD-Net obtains a score of 84.1\% in terms of Rank-1 accuracy and 63.7\% in terms of mAP on MSMT17. In short, the results on MSMT17 are consistent with those for the first three databases.

\section{Conclusion}
In this work, we propose a novel framework, named BCD-Net, which explores batch-level statistics in order to drive the ReID model to extract diverse and semantically aligned part-level features. Our contributions are twofold: i.e., we present a batch coherence-guided channel attention (BCCA) module and a pair of spatial regularization terms. BCCA highlights the relevant channels for each respective part. The supervision signals for part-channel correspondence are summarized from each batch of training images. Moreover, the part-level spatial regularization supervises BCD-Net, causing it to emphasize disjoint part-level spatial regions and learn diverse part-level representations, while the holistic-level spatial regularization is designed to explore the complementary information pertaining to body parts. Furthermore, unlike most existing part-aware ReID works, BCD-Net bypasses the part detection step entirely during both the training and testing phases. We conduct extensive experiments on four large-scale ReID benchmarks, thereby demonstrating the effectiveness of the proposed model and its ability to achieve state-of-the-art performance.

\bibliographystyle{IEEEtran}
\bibliography{egbib}

%

\begin{IEEEbiography}[{\includegraphics[width=1in,height=1.25in,clip,keepaspectratio]{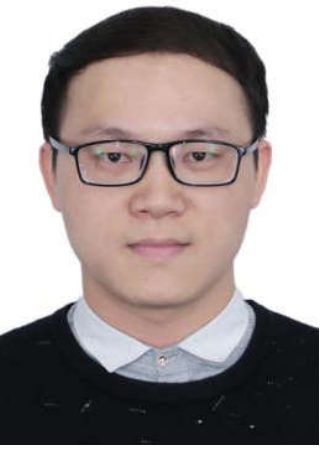}}]{Kan Wang}
received the B.S. degree in mathematics from South China University of Technology, Guangzhou, China, in 2016, where he is currently pursuing the Ph.D. degree with the School of Electronic and Information Engineering. His
current research interests include computer vision, deep learning, and person re-identification.
\end{IEEEbiography}

\begin{IEEEbiography}[{\includegraphics[width=1in,height=1.25in,clip,keepaspectratio]{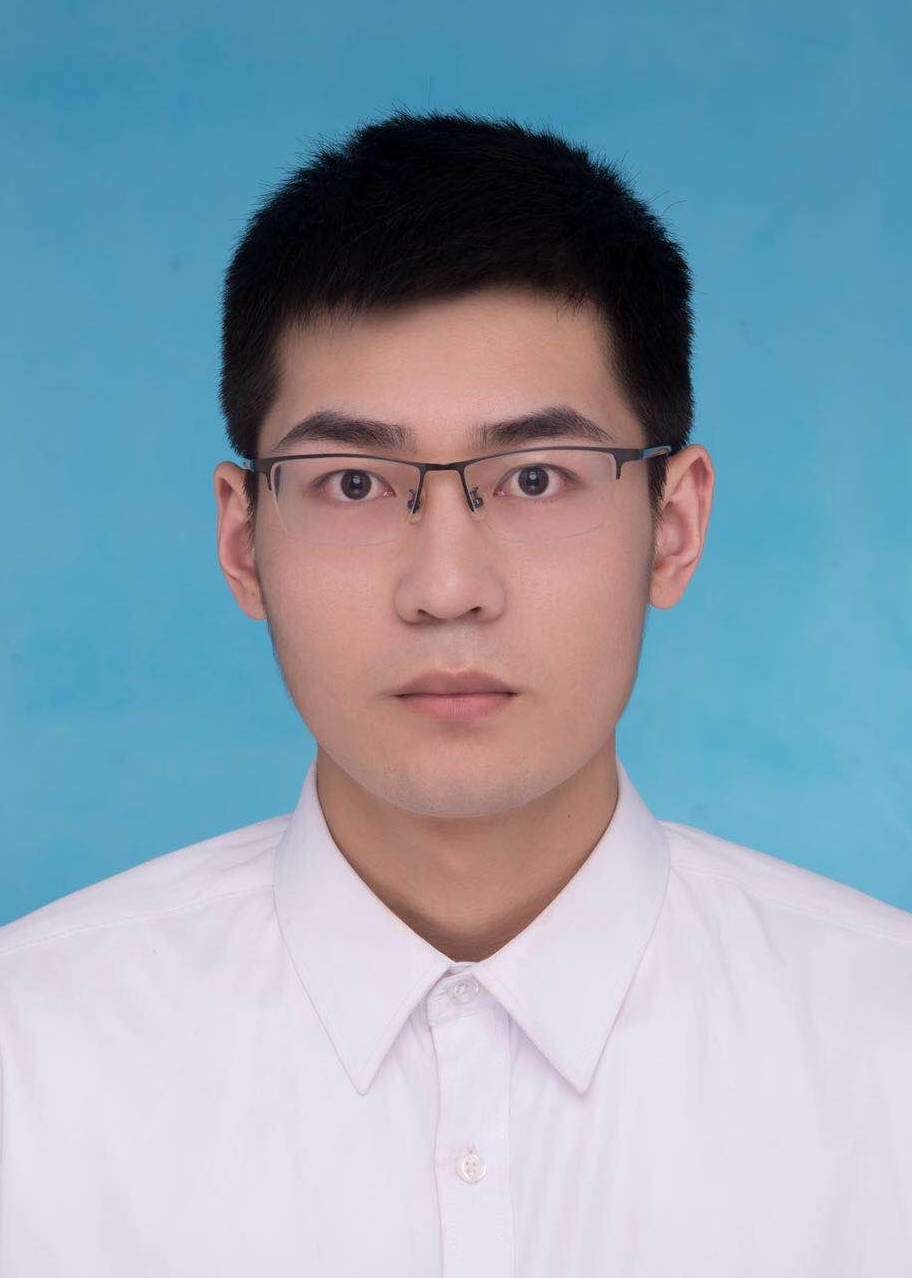}}]{Pengfei Wang}
received the B.S. degree from South China University of Technology, Guangzhou, China, in 2019, where he is currently working towards the M.S. degree with the School of Electronic and Information Engineering. His current research interests include computer vision, deep learning, and person re-identification.
\end{IEEEbiography}

\begin{IEEEbiography}[{\includegraphics[width=1in,height=1.25in,clip,keepaspectratio]{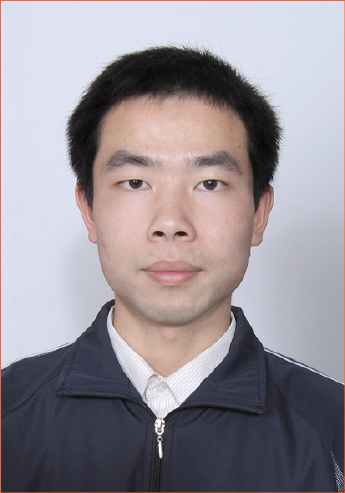}}]{Changxing Ding} (M'16) received the Ph.D. degree from the University of Technology Sydney, Australia, in 2016. He is now a research associate professor with the School of Electronic and Information Engineering, South China University of Technology. He has published 20 papers in prominent journals and conferences, including the IEEE T-PAMI, IEEE T-IP, IEEE T-MM, CVPR, ECCV, and MICCAI, etc. His research interests include visual relationship detection, person re-identification, and face recognition.
\end{IEEEbiography}

\begin{IEEEbiography}[{\includegraphics[width=1in,height=1.25in,clip,keepaspectratio]{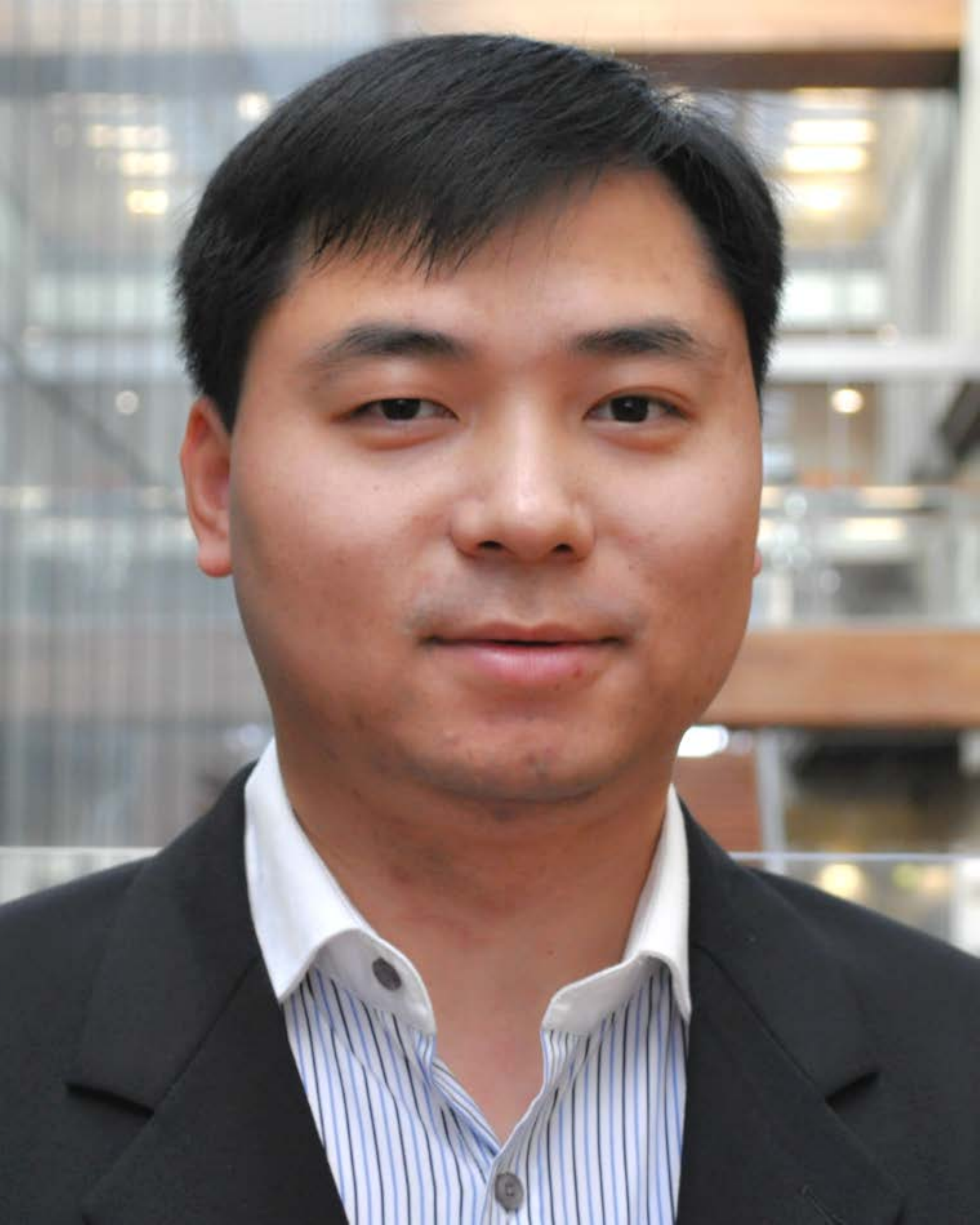}}]{Dacheng Tao} (F'15) is currently the director of the JD Explore Academy and VP at JD.com. He mainly applies statistics and mathematics to artificial intelligence and data science, and his research is detailed in one monograph and over 200 publications in prestigious journals and proceedings at leading conferences. He received the 2015 Australian Scopus-Eureka Prize, the 2018 IEEE ICDM Research Contributions Award, and the 2021 IEEE Computer Society McCluskey Technical Achievement Award. He is a fellow of the Australian Academy of Science, AAAS, ACM and IEEE.
\end{IEEEbiography}







\end{document}